\documentclass[runningheads]{llncs}

\usepackage[mobile]{eccv}

\usepackage{eccvabbrv}

\usepackage{graphicx}
\usepackage{booktabs}

\usepackage[accsupp]{axessibility}  %

\usepackage[pagebackref,breaklinks,colorlinks,citecolor=eccvblue]{hyperref}

\usepackage{orcidlink}

\usepackage{graphicx}
\usepackage{amsmath}
\usepackage{amssymb}
\usepackage{booktabs}
\usepackage{lipsum}
\usepackage[dvipsnames]{xcolor}
\usepackage{physics}
\usepackage{dsfont}
\usepackage{multirow}
\usepackage{balance}
\usepackage{duckuments}
\usepackage{capt-of}
\usepackage{arydshln}
\usepackage[pagebackref,breaklinks,colorlinks]{hyperref}

\PassOptionsToPackage{sort,comma,numbers}{natbib}
\usepackage[capitalize]{cleveref}
\crefname{section}{Sec.}{Secs.}
\Crefname{section}{Section}{Sections}
\Crefname{table}{Table}{Tables}
\crefname{table}{Tab.}{Tabs.}

\begin{document}

\title{Distilling Diffusion Models into \\ Conditional GANs}
\author{}
\author{Minguk Kang\inst{1, 2} \and
Richard Zhang\inst{2} \and
Connelly Barnes\inst{2} \and
\\ Sylvain Paris\inst{2} \and
Suha Kwak\inst{1} \and
Jaesik Park\inst{3}\and
\\ Eli Shechtman\inst{2} \and
Jun-Yan Zhu\inst{4} \and
Taesung Park\inst{2}
}

\institute{}
\institute{Pohang University of Science and Technology\inst{1}\; \;
Adobe Research\inst{2}\; \;  \\
Seoul National University\inst{3}\; \;
Carnegie Mellon University\inst{4}\; \;}

\authorrunning{Kang et al.}

\maketitle

\newcommand{\minguk}[1]{{\textcolor{red}{#1}}}
\newcommand{\jaesik}[1]{{\textcolor{violet}{Jaesik: #1}}}
\newcommand{\tae}[1]{{\textcolor{red}{Taesung: #1}}}
\newcommand{\connelly}[1]{{\textcolor{orange}{Connelly: #1}}}
\newcommand{\richard}[1]{{\textcolor{blue}{Rich: #1}}}

\definecolor{darkblue}{HTML}{202282} 
\definecolor{lightgray}{HTML}{E1E1E1} 
\newcommand{\newtext}[1]{{\textcolor{violet}{#1}}}

\newcommand{\perm}{\boldsymbol{\pi}}
\newcommand{\myparagraph}[1]{\vspace{-2pt} \smallskip \noindent \textbf{#1}}

\newcommand{\mysubsection}[1]{\vspace{1mm}\noindent{\bf #1}}

\newcommand{\cmark}{\ding{51}}
\newcommand{\xmark}{\ding{55}}

\def\noise{\rvz}
\def\image{\rvx}
\def\prompt{\vb{c}}
\def\L{\mathcal{L}}

\def\vpos{\bm{v^{+}}}
\def\vneg{\bm{v^{-}}}

\def\X{\mathcal{X}}
\def\Y{\mathcal{Y}}
\def\Z{\mathcal{Z}}
\def\data{\mathcal{D}}
\def\P{\mathbb{P}}
\def\Px{\mathbb{P}_{\mathcal{X}}}
\def\Pg{\mathbb{P}_{\theta}}
\def \Lv{\mathcal{L}_{\text{models}}}
\def \kmax{K\max}

\def\xhat{{\hat{\bm x}}}
\def\yhat{{\hat{\bm y}}}
\def\zhat{{\hat{\bm z}}}

\def\xtilde{{\tilde{\x}}}
\def\ytilde{{\tilde{\y}}}
\newcommand*\wc{{\mkern 2mu\cdot\mkern 2mu}}

\def\ztilde{{\tilde{\z}}}

\def\xbar{{\bar{\x}}}
\def\ybar{{\bar{\y}}}
\def\zbar{{\bar{\z}}}

\def\xstar{{\x^{*}}}
\def\ystar{{\y^{*}}}
\def\zstar{{\z^{*}}}
\def\dist{d}

\def\Re{\mathds{R}}
\newcommand{\expect}[1]{\mathbb{E}_{#1}}
\def\NCE{\ell}

\newcommand{\citeColored}[2]{{\hypersetup{citecolor=#1}\cite{#2}}}

\newcommand{\fid}{Fr\'echet Inception Distance\xspace}

\def\rvepsilon{{\mathbf{\epsilon}}}
\def\rvtheta{{\mathbf{\theta}}}
\def\rva{{\mathbf{a}}}
\def\rvb{{\mathbf{b}}}
\def\rvc{{\mathbf{c}}}
\def\rvd{{\mathbf{d}}}
\def\rve{{\mathbf{e}}}
\def\rvf{{\mathbf{f}}}
\def\rvg{{\mathbf{g}}}
\def\rvh{{\mathbf{h}}}
\def\rvu{{\mathbf{i}}}
\def\rvj{{\mathbf{j}}}
\def\rvk{{\mathbf{k}}}
\def\rvl{{\mathbf{l}}}
\def\rvm{{\mathbf{m}}}
\def\rvn{{\mathbf{n}}}
\def\rvo{{\mathbf{o}}}
\def\rvp{{\mathbf{p}}}
\def\rvq{{\mathbf{q}}}
\def\rvr{{\mathbf{r}}}
\def\rvs{{\mathbf{s}}}
\def\rvt{{\mathbf{t}}}
\def\rvu{{\mathbf{u}}}
\def\rvv{{\mathbf{v}}}
\def\rvw{{\mathbf{w}}}
\def\rvx{{\mathbf{x}}}
\def\rvy{{\mathbf{y}}}
\def\rvz{{\mathbf{z}}}

\newcommand{\seg}{\rvs_c}
\newcommand{\repr}{\rvr}
\newcommand{\repru}{\repr_{u, \sP}}
\newcommand{\uprepru}{\repr_{u, \sP}^{\uparrow}}
\newcommand{\f}{f}
\newcommand{\h}{h}

\newcommand{\figleft}{{\em (Left)}}
\newcommand{\figcenter}{{\em (Center)}}
\newcommand{\figright}{{\em (Right)}}
\newcommand{\figtop}{{\em (Top)}}
\newcommand{\figbottom}{{\em (Bottom)}}
\newcommand{\captiona}{{\em (a)}}
\newcommand{\captionb}{{\em (b)}}
\newcommand{\captionc}{{\em (c)}}
\newcommand{\captiond}{{\em (d)}}

\newcommand{\newterm}[1]{{\bf #1}}

\newcommand{\reffig}[1]{Figure~\ref{fig:#1}}
\newcommand{\refsec}[1]{Section~\ref{sec:#1}}
\newcommand{\refapp}[1]{Appendix~\ref{sec:#1}}
\newcommand{\reftbl}[1]{Table~\ref{tbl:#1}}
\newcommand{\refalg}[1]{Algorithm~\ref{alg:#1}}
\newcommand{\refline}[1]{Line~\ref{line:#1}}
\newcommand{\shortrefsec}[1]{\S~\ref{sec:#1}}
\newcommand{\refeq}[1]{Eqn.~\ref{eq:#1}}
\newcommand{\refeqshort}[1]{(\ref{eq:#1})}
\newcommand{\shortrefeq}[1]{\ref{eq:#1}}
\newcommand{\lblfig}[1]{\label{fig:#1}}
\newcommand{\lblsec}[1]{\label{sec:#1}}
\newcommand{\lbleq}[1]{\label{eq:#1}}
\newcommand{\lbltbl}[1]{\label{tbl:#1}}
\newcommand{\lblalg}[1]{\label{alg:#1}}
\newcommand{\lblline}[1]{\label{line:#1}}
\newcommand{\ignorethis}[1]{}
\newcommand{\revision}[1]{\color{black}#1\color{black}}
\newcommand{\myitem}{\vspace{-5pt}\item}

\def\ceil#1{\lceil #1 \rceil}
\def\floor#1{\lfloor #1 \rfloor}
\def\1{\bm{1}}
\newcommand{\train}{\mathcal{D}}
\newcommand{\valid}{\mathcal{D_{\mathrm{valid}}}}
\newcommand{\test}{\mathcal{D_{\mathrm{test}}}}

\def\eps{{\epsilon}}

\newcommand{\images}{{\mathcal{X}}}
\newcommand{\imagedist}{{p_{\text{data}}(\image)}}
\newcommand{\latentdist}{{p(\latent)}}
\newcommand{\imageD}{{D_X}}
\newcommand{\Fnet}{{F}}
\newcommand{\sketch}{{\rvy}}
\newcommand{\sketches}{{\mathcal{Y}}}
\newcommand{\sketchdist}{{p_{\text{data}}(\sketch)}}
\newcommand{\sketchD}{{D_Y}}
\newcommand{\modelold}{{G(\rvz; \theta)}}
\newcommand{\modelnew}{{G(\rvz; \theta')}}
\newcommand{\losssketch}{{\mathcal{L}_{\text{sketch}}}}
\newcommand{\lossimage}{{\mathcal{L}_{\text{image}}}}
\newcommand{\lossweight}{{\mathcal{L}_{\text{weight}}}}

\newcommand{\method}{{GAN Sketching}}

\newcommand{\pdata}{{D}}
\newcommand{\ptrain}{\hat{p}_{\rm{data}}}
\newcommand{\Ptrain}{\hat{P}_{\rm{data}}}
\newcommand{\pmodel}{p_{\rm{model}}}
\newcommand{\Pmodel}{P_{\rm{model}}}
\newcommand{\ptildemodel}{\tilde{p}_{\rm{model}}}
\newcommand{\pencode}{p_{\rm{encoder}}}
\newcommand{\pdecode}{p_{\rm{decoder}}}
\newcommand{\precons}{p_{\rm{reconstruct}}}

\newcommand{\laplace}{\mathrm{Laplace}} %

\newcommand{\Ls}{\mathcal{L}}
\newcommand{\R}{\mathbb{R}}
\newcommand{\emp}{\tilde{p}}
\newcommand{\lr}{\alpha}
\newcommand{\reg}{\lambda}
\newcommand{\rect}{\mathrm{rectifier}}
\newcommand{\softmax}{\mathrm{softmax}}
\newcommand{\sigmoid}{\sigma}
\newcommand{\softplus}{\zeta}
\newcommand{\Var}{\mathrm{Var}}
\newcommand{\standarderror}{\mathrm{SE}}
\newcommand{\Cov}{\mathrm{Cov}}
\newcommand{\normlzero}{L^0}
\newcommand{\normlone}{L^1}
\newcommand{\normltwo}{L^2}
\newcommand{\normlp}{L^p}
\newcommand{\normmax}{L^\infty}

\newcommand{\parents}{Pa} %

\newcommand{\xpar}[1]{\noindent\textbf{#1}\ \ }
\newcommand{\vpar}[1]{\vspace{3mm}\noindent\textbf{#1}\ \ }

\newcommand{\shapenet}{ShapeNet\xspace}
\newcommand{\pascal}{PASCAL 3D+\xspace}

\newcommand{\degree}{\ensuremath{^\circ}\xspace}
\newcommand{\ignore}[1]{}

\newcommand{\fcseven}{$\mbox{fc}_7$}

\renewcommand*{\thefootnote}{\arabic{footnote}}

\def\naive{na\"{\i}ve\xspace}
\def\Naive{Na\"{\i}ve\xspace}

\makeatletter
\DeclareRobustCommand\onedot{\futurelet\@let@token\@onedot}
\def\@onedot{\ifx\@let@token.\else.\null\fi\xspace}
\def\eg{e.g\onedot,\xspace} 
\def\Eg{E.g\onedot,}
\def\ie{i.e\onedot,\xspace} 
\def\Ie{\emph{I.e}\onedot,}
\def\cf{\emph{c.f}\onedot} \def\Cf{\emph{C.f}\onedot}
\def\etc{\emph{etc}\onedot} \def\vs{\emph{vs}\onedot}
\def\wrt{w.r.t\onedot} \def\dof{d.o.f\onedot}
\def\etal{\emph{et al}\onedot}
\makeatother

\newcommand*{\img}[1]{%
    \raisebox{-.25\baselineskip}{%
        \includegraphics[
        height=\baselineskip,
        width=\baselineskip,
        keepaspectratio,
        ]{#1}%
    }%
}

\fboxsep=0mm%
\fboxrule=2pt%
\newcommand*{\myoverpic}[2]{
  \begin{overpic}[width=.18\linewidth]{#1}
     \put(60,63){\fbox{\includegraphics[width=.06\linewidth]{#2}}}  
  \end{overpic}
}

\newcommand{\OursAcronym}{Diffusion2GAN}

\newcommand{\figwidth}{0.98\linewidth}

\vspace{-3mm}

\begin{abstract}
We propose a method to distill a complex multistep diffusion model into a single-step conditional GAN student model, dramatically accelerating inference, while preserving image quality.
Our approach interprets diffusion distillation as a \textit{paired image-to-image translation} task, using noise-to-image pairs of the diffusion model’s ODE trajectory. For efficient regression loss computation, we propose E-LatentLPIPS, a perceptual loss operating directly in diffusion model's latent space, utilizing an ensemble of augmentations.  
Furthermore, we adapt a diffusion model to construct a multi-scale discriminator with a text alignment loss to build an effective conditional GAN-based formulation. E-LatentLPIPS converges more efficiently than many existing distillation methods, even accounting for dataset construction costs.
We demonstrate that our one-step generator outperforms cutting-edge one-step diffusion distillation models -- DMD, SDXL-Turbo, and SDXL-Lightning -- on the zero-shot COCO benchmark.
\end{abstract}

\section{Introduction}
Diffusion models~\cite{sohl2015deep, Ho2020DenoisingDP, song2021scorebased} have demonstrated unprecedented image synthesis quality on challenging datasets, such as LAION~\cite{schuhmann2022laion}. However, producing high-quality results requires dozens or hundreds of sampling steps. As a result, most existing diffusion-based image generation models, such as DALL$\cdot$E~2~\cite{ramesh2022hierarchical}, Imagen~\cite{saharia2022photorealistic}, and Stable Diffusion~\cite{rombach2022high}, suffer from high latency, often exceeding 10 seconds and hindering real-time interaction. If our model only requires \emph{one} inference step, it will not only improve the user experience in text-to-image synthesis, but also expand its potential in 3D and video applications~\cite{poole2022dreamfusion,ho2022imagen}.
But how can we build a one-step text-to-image model?

One simple solution is to just train a one-step model from scratch. For example, we can train a GAN~\cite{Goodfellow2014GAN}, a leading one-step model for simple domains~\cite{karras2019style}. Unfortunately, training 
text-to-image GANs on large-scale and diverse datasets is still challenging, despite recent advances~\cite{kang2023scaling,Sauer2023StyleGANTUT}. 
The challenge lies in GANs needing to tackle \emph{two} difficult tasks all at once without any supervision: (1) finding correspondence between noises and natural images, and (2) effectively optimizing a generator model to perform the mapping from noises to images. This ``unpaired'' learning is often considered more ill-posed, as mentioned in CycleGAN~\cite{zhu2017unpaired}, compared to paired learning, where conditional GANs~\cite{isola2017image} can learn to map the input to output, given ground truth correspondences. 

Our key idea is to tackle the above tasks one by one. We first find the correspondence between noises and images by simulating the ODE solver with a pre-trained diffusion model. Given the established corresponding pairs, we then ask a conditional GAN to map noises to images in a paired image-to-image translation framework~\cite{isola2017image,park2019semantic}.  This disentangled approach allows us to leverage two types of generative models for separate tasks, achieving the benefits of both: finding high-quality correspondence using diffusion models, while achieving fast mapping using conditional GANs.

In this work, we collect a large number of noise-to-image pairs from a pre-trained diffusion model and treat the task as a paired image-to-image translation problem~\cite{isola2017image}, enabling us to exploit tools such as perceptual losses~\cite{johnson2016perceptual,dosovitskiy2016generating,zhang2018unreasonable} and conditional GANs~\cite{Goodfellow2014GAN,Mirza2014ConditionalGA,isola2017image}. In doing so, we make a somewhat unexpected discovery. Collecting a large database of noise-image pairs and training with a regression loss without the GAN loss can already achieve comparable results to more recent distillation methods~\cite{song2023consistency,meng2023distillation}, at a significantly lower compute budget, if the regression loss is designed carefully. 

First, in regression tasks, using perceptual losses (such as LPIPS~\cite{zhang2018unreasonable}) better preserves perceptually important details over point-based losses (such as L2).
However, perceptual losses are fundamentally incompatible with Latent Diffusion Models~\cite{rombach2022high}, as they require an expensive decoding from latent to pixel space. To overcome this, we propose LatentLPIPS, showing that perceptual losses can directly work in latent space. This enables a fourfold increase in batch size, compared to computing LPIPS in pixel space. Unfortunately, we observe that the latent-based perceptual loss has more blind spots than its pixel counterparts. While previous work has found that ensembling is helpful for pixel-based LPIPS~\cite{kettunen2019lpips}, we find that it is critical for the latent-based version. Working in latent space with our Ensembled-LatentLPIPS, we demonstrate strong performance with just a regression loss, comparable to guided progressive distillation~\cite{meng2023distillation}. Additionally, we employ a discriminator in the training to further improve performance. We develop a multi-scale conditional diffusion discriminator, leveraging the pre-trained weights and using our new single-sample R1 loss and mix-and-match augmentation. We name our distillation model \textit{\OursAcronym{}.}

Using the proposed \OursAcronym{} framework, we distill Stable Diffusion 1.5~\cite{rombach2022high} into a single-step conditional GAN model.
Our Diffusion2GAN can learn noise-to-image correspondences inherent in the target diffusion model better than other distillation methods. It also outperforms recently proposed distillation models, UFOGen~\cite{xu2023ufogen} and DMD~\cite{yin2023onestep}, on the zero-shot one-step COCO2014~\cite{lin2014microsoft} benchmark.
Furthermore, we perform extensive ablation studies and highlight the critical roles of both E-LatentLPIPS and multi-scale diffusion discriminator. Beyond the distillation of Stable Diffusion 1.5, we demonstrate the effectiveness of \OursAcronym{} in distilling a larger SDXL~\cite{podell2024sdxl}, exhibiting superior FID~\cite{Heusel2017GANsTB} and CLIP-score~\cite{hessel2021clipscore} over one-step SDXL-Turbo~\cite{sauer2023adversarial} and SDXL-Lightning~\cite{lin2024sdxl}.

\begin{figure*}[htp!]
    \centering
    \includegraphics[width=1.0\linewidth]{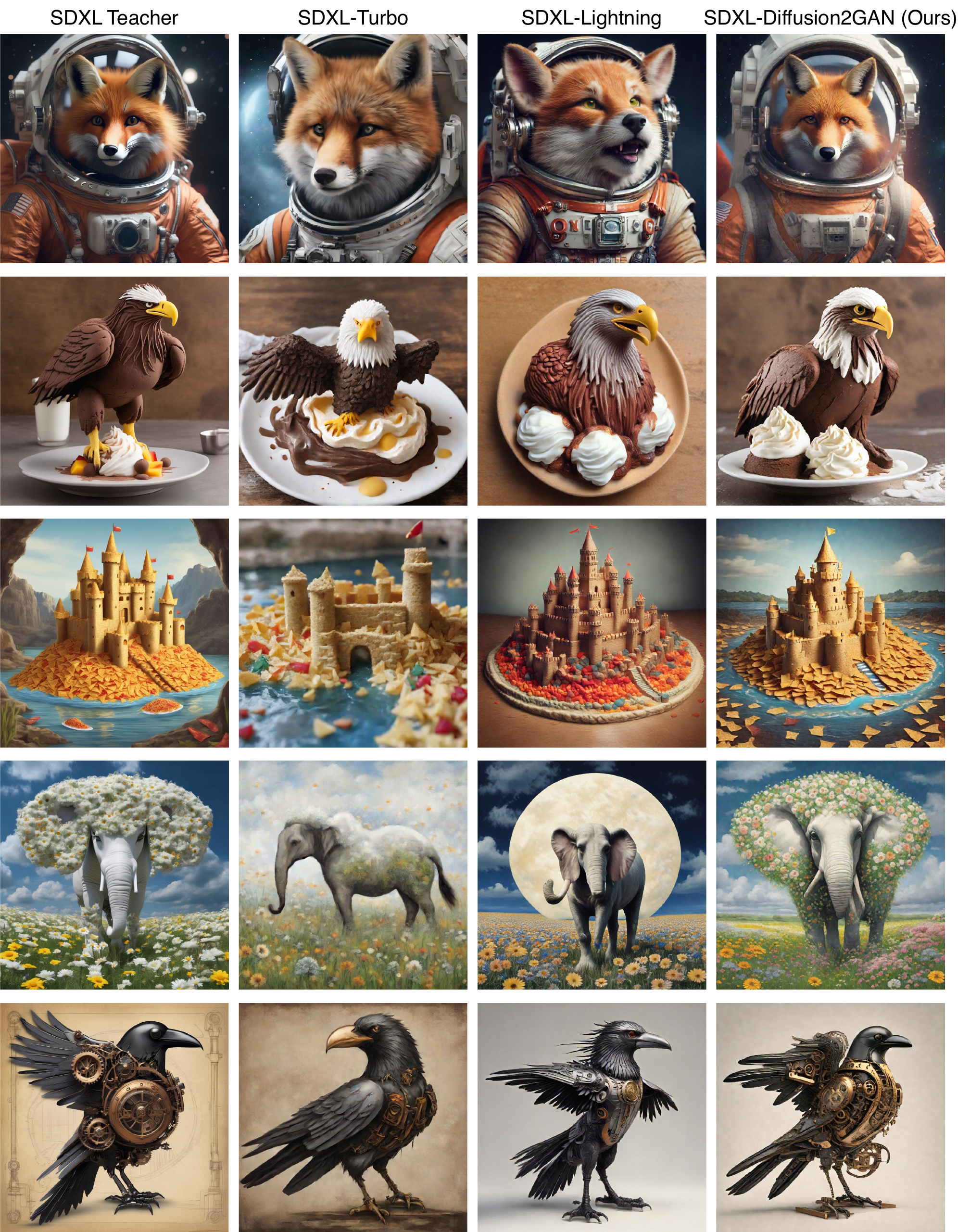}
    \captionsetup{width=1.0\linewidth}
    \captionsetup{singlelinecheck = false, justification=justified}
    \caption{Visual comparison to SDXL teacher~\cite{podell2024sdxl}  with a classifier-free guidance scale~\cite{ho2022classifier} of 7 and selected distillation student models, including SDXL-Turbo~\cite{sauer2023adversarial}, SDXL-Lightning~\cite{lin2024sdxl}, and our SDXL-\OursAcronym{}. All images in a given row were generated using the same noise input, except for SDXL-Turbo, which requires a distinct noise size of $4\times64\times64$.
    Compared to other distillation models, our SDXL-Diffusion2GAN more closely adheres to the original ODE trajectory. }
    \label{fig:visual_comparison2}
\end{figure*}

\begin{figure*}[htp!]
    \centering
    \includegraphics[width=1.0\linewidth]{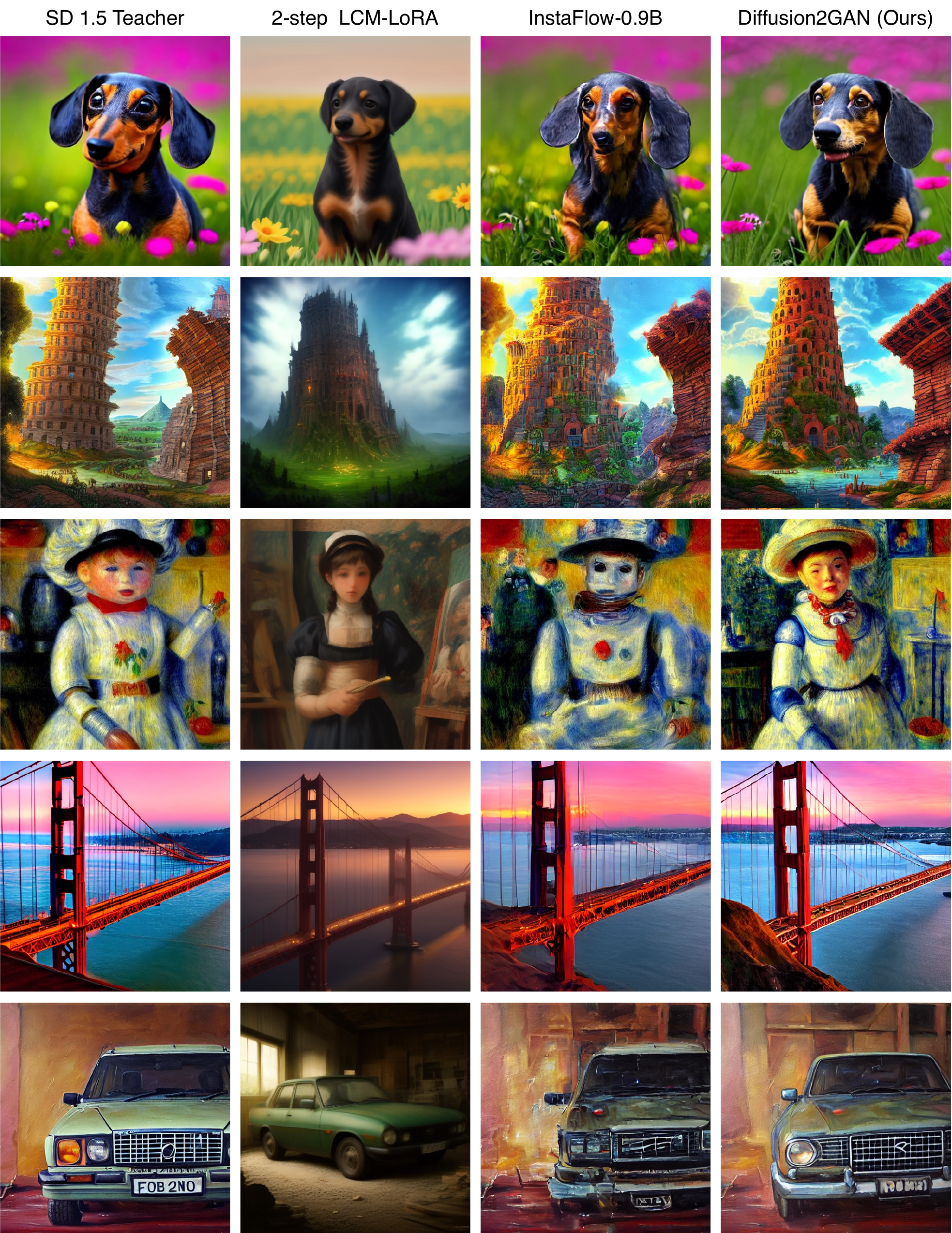}
    \captionsetup{width=1.0\linewidth}
    \captionsetup{singlelinecheck = false, justification=justified}
    \caption{Visual comparison to Stable Diffusion 1.5 teacher~\cite{stablediffusion1.5} with a classifier-free guidance scale~\cite{ho2022classifier} of 8 and selected distillation student models, including InstaFlow-0.9B~\cite{liu2023insta}, LCM-LoRA~\cite{luo2023latentlora}, and our \OursAcronym{}. The same noise input was used to generate images in the same row. 
    Our method Diffusion2GAN achieves higher realism than the 2-step LCM-LoRA and InstaFlow-0.9B.}
    \label{fig:visual_comparison}
\end{figure*}

\begin{figure*}[htp!]
    \centering
    \includegraphics[width=0.94\linewidth]{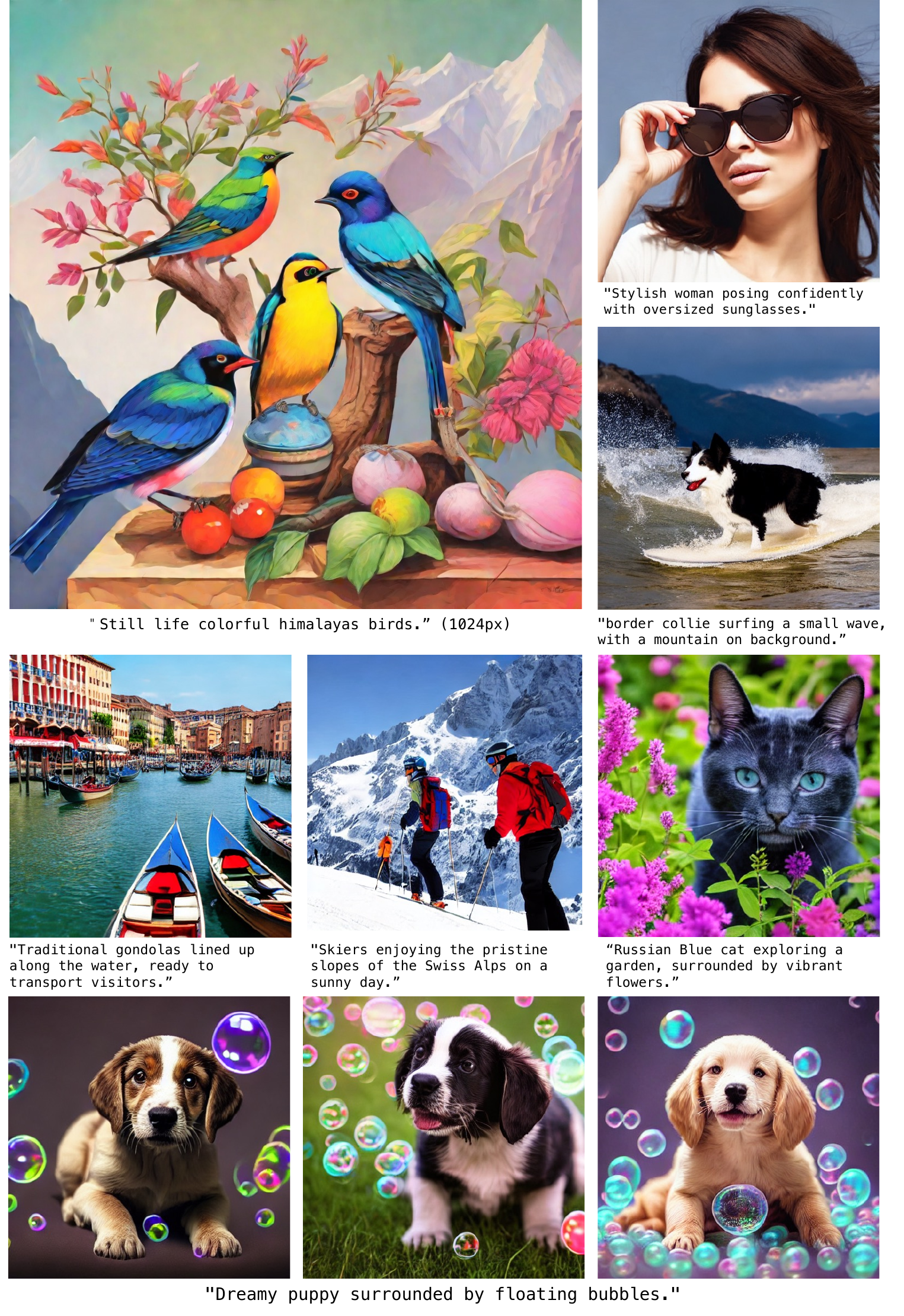}
    \vspace{-3mm}
    \captionsetup{width=1.0\linewidth}
    \captionsetup{singlelinecheck = false, justification=justified}    
    \caption{High-quality generated images using our one-step \OursAcronym{} framework. Our model can synthesize a 512px/1024px image at an interactive speed of 0.09/0.16 seconds on an A100 GPU, while the teacher model, Stable Diffusion 1.5~\cite{stablediffusion1.5}/SDXL~\cite{podell2024sdxl}, produces an image in 2.59/5.60 seconds using 50 steps of the DDIM~\cite{song2021denoising}.  Please visit our \href{https://mingukkang.github.io/Diffusion2GAN/}{website} for more visual results.}
    \label{fig:text2image_results}
\end{figure*}

\section{Related Work}
\myparagraph{Diffusion models.} Diffusion models (DMs)~\cite{sohl2015deep, Ho2020DenoisingDP, song2021scorebased} are a family of generative models consisting of the diffusion process and denoising process. 
The diffusion process progressively diffuses high-dimensional data from data distribution to easy-to-sample Gaussian distribution, while the denoising process aims to reverse the process using a deep neural network trained on a score-matching objective~\cite{vincent2011connection, song2020sliced, song2021scorebased}. Once trained, these models can generate data from random Gaussian noise, using numerical integrators~\cite{atkinson1991introduction, ascher1998computer, karras2022elucidating}. Diffusion models have enabled numerous vision and graphics applications, such as image editing~\cite{meng2022sdedit,hertz2022prompt,brooks2023instructpix2pix}, controllable image synthesis~\cite{zhang2023adding,mou2023t2i}, personalized generation~\cite{ruiz2023dreambooth,gal2022image,kumari2023multi}, video synthesis~\cite{ho2022imagen,blattmann2023stable,guo2023animatediff}, and 3D content creation~\cite{poole2022dreamfusion,lin2023magic3d}. However, the sampling typically requires tens of sampling steps, leading to slower image generation speed than other generative models, such as GANs~\cite{Goodfellow2014GAN} and VAEs~\cite{kingma2013auto}. In this work, our goal is to accelerate the model's inference while maintaining image quality. %

\myparagraph{Diffusion distillation.} Accelerating the sampling speed of diffusion models is crucial for enhancing practical applications, as well as reducing energy costs for inference. Several works have proposed reducing the number of sampling steps using fast ODE solvers~\cite{lu2022dpm,lu2022dpm++,karras2022elucidating} or reducing the computational time per step~\cite{li2022efficient,li2024snapfusion,chen2023speed}. Another effective method for acceleration is knowledge distillation~\cite{Hinton2015DistillingTK, luhman2021knowledge, xiao2022DDGAN, liu2022flow, salimans2022progressive, meng2023distillation, berthelot2023tract, gu2023boot, zheng2023fast, liu2023insta, song2023consistency}. In this approach, multiple steps of a teacher diffusion model is distilled into a fewer-step student model. Luhman~\etal~\cite{luhman2021knowledge} propose $L_{p}$ regression training between student's output from a Gaussian noise $\vb{x}_{T}$ and its corresponding ODE solution $\vb{x}_{0}$. Despite its simplicity, such direct regression produces blurry outputs and does not match the image synthesis capabilities exhibited by other generative models. To enhance image quality, InstaFlow~\cite{liu2023insta} straightens high-curvature ODE trajectory via ReFlow~\cite{liu2022flow} and distills the linearized ODE trajectory to the student model. Consistency Distillation~(CD)~\cite{song2023consistency,luo2023latent} trains a student model to predict a consistent output for any noisy sample $\vb{x}_{t+1}$ and its single-step denoising $\vb{x}_{t}$. Recently, several studies have proposed using a GAN discriminator to enhance distillation performance. For example, CTM~\cite{kim2023consistency} and SDXL-Turbo~\cite{sauer2023adversarial} utilize an improved StyleGAN~\cite{sauer2022stylegan, Sauer2023StyleGANTUT} discriminator to train a one-step image generator. In addition, UFOGen~\cite{xu2023ufogen}, SDXL-Lightning~\cite{lin2024sdxl}, and LADD~\cite{sauer2024fast} adopt a pre-trained diffusion model as a strong discriminator, demonstrating their capabilities in one-step text-to-image synthesis. Although these works are concurrent, we will compare our method with SDXL-Turbo and SDXL-Lightning, both of which have demonstrated state-of-the-art performance in diffusion distillation for one-step image synthesis.

\myparagraph{Conditional Generative Adversarial Networks}~\cite{Mirza2014ConditionalGA,isola2017image} have been a commonly-used framework for conditional image synthesis. The condition could be an image~\cite{isola2017image, zhu2017unpaired, Choi_2018_CVPR, Liu_2019_ICCV, park2020cut, Richardson_2021_CVPR}, class-label~\cite{Odena2017ConditionalIS, Miyato2018cGANsWP, Brock2019LargeSG, karras2020analyzing, Kang2021RebootingAA}, and text~\cite{reed2016learning, zhang2017stackgan, xu2018attngan, Sauer2023StyleGANTUT, kang2023scaling}.
In particular, cGANs have shown impressive performance when helped by a regression loss to stabilize training, as in image translation~\cite{isola2017image,Wang2018HighResolutionIS,zhu2017unpaired,park2019semantic,zhao2021comodgan,park2020swapping}. Likewise, we approach diffusion model distillation by employing the image-conditional GAN, along with a perceptual regression loss~\cite{zhang2018unreasonable}. %
Early works~\cite{xiao2022DDGAN, wang2023diffusiongan} combine GANs with the forward diffusion process, but do not aim at distilling a pre-trained diffusion model into a GAN.

\section{Method}
\label{sec:method}
Our goal is to distill a pre-trained text-to-image diffusion model into a one-step generator. That is, we want to learn a mapping $\image = G(\noise, \prompt)$, with one-step generator network $G$ mapping input noise $\noise$ and text $\prompt$ to the diffusion model output $\image$. We assume that the student and teacher share the same architecture, so that we can initialize the student model $G$ using weights of the teacher model. For our method section, we assume Latent Diffusion Models~\cite{rombach2022high} with $\image, \noise \in \mathbb{R}^{4 \times 64 \times 64}$. Later, we also adopt our method to the SDXL model~\cite{podell2024sdxl}. 

In the rest of this section, we will elaborate on the design and training principles of our framework. 
We begin by describing distillation as a paired image-to-image translation problem in Section~\ref{sec:generator}.
Then, we introduce our Ensembled Latent LPIPS regression loss (E-LatentLPIPS) in Section~\ref{sec:elatentlpips}. Just using this regression loss improves training efficiency and significantly improves distillation performance for latent diffusion models. Lastly, we present an improved discriminator design that reuses a pre-trained diffusion model~(Section~\ref{sec:discriminator}). It is worth noting that our findings extend beyond the specific type of latent space diffusion models~\cite{rombach2022high, stablediffusion} and apply to a pixel space model~\cite{karras2022elucidating} as well.

\subsection{Paired Noise-to-Image Translation for One-step Generation}
\label{sec:generator}
With the emergence of diffusion probabilistic models~\cite{Ho2020DenoisingDP, song2021scorebased}, Luhman~\etal~\cite{luhman2021knowledge} suggest that the multi-step denoising process of a pre-trained diffusion model can be reduced to a single step by minimizing the following distillation objective:
\begin{equation}
  \mathcal{L}_{\text{distill}}^{\text{ODE}} = \mathbb{E}_{\{\noise, \prompt, \image\}}\Big[ \dist(G( \noise, \prompt), \image)\Big],
  \label{eq:our_regression_loss}
\end{equation}
where $\noise$ is a sample from Gaussian noise, $\prompt$ is a text prompt, $G$ denotes a UNet generator with trainable weights, $\image$ is the output of the diffusion model simulating the ordinary differential equation~(ODE) trajectory with the DDIM sampler~\cite{song2021denoising}, and $\dist(\cdot, \cdot)$ is a distance metric. Due to the computational cost of obtaining $\image$ for each iteration, the method uses pre-computed pairs of noise and corresponding ODE solutions before training begins. During training, it randomly samples noise-image pairs and minimizes the ODE distillation loss (Equation~\ref{eq:our_regression_loss}). While the proposed approach looks simple and straightforward, the direct distillation approach yields inferior image synthesis results compared to more recent distillation methods~\cite{salimans2022progressive, meng2023distillation, song2023consistency, liu2023insta}. 

In our work, we hypothesize that the full potential of direct distillation has not yet been realized. In our experiments on CIFAR10, we observe that we can significantly improve the quality of distillation by (1) scaling up the size of the ODE pair dataset and (2) using a perceptual loss~\cite{zhang2018unreasonable} (as opposed to the pixel-space L2 loss in Luhman~\etal). In Table~\ref{table:cifar10_exp}, we show the training progression on the CIFAR10 dataset, and compare its performance to Consistency Model~\cite{song2023consistency}. Surprisingly, the direct distillation with the LPIPS loss can achieve lower FID than the Consistency Model at smaller amount of total compute, even accounting for the extra compute to collect the ODE pairs.

\begin{figure*}[t!]
    \centering
    \includegraphics[width=0.98\linewidth]{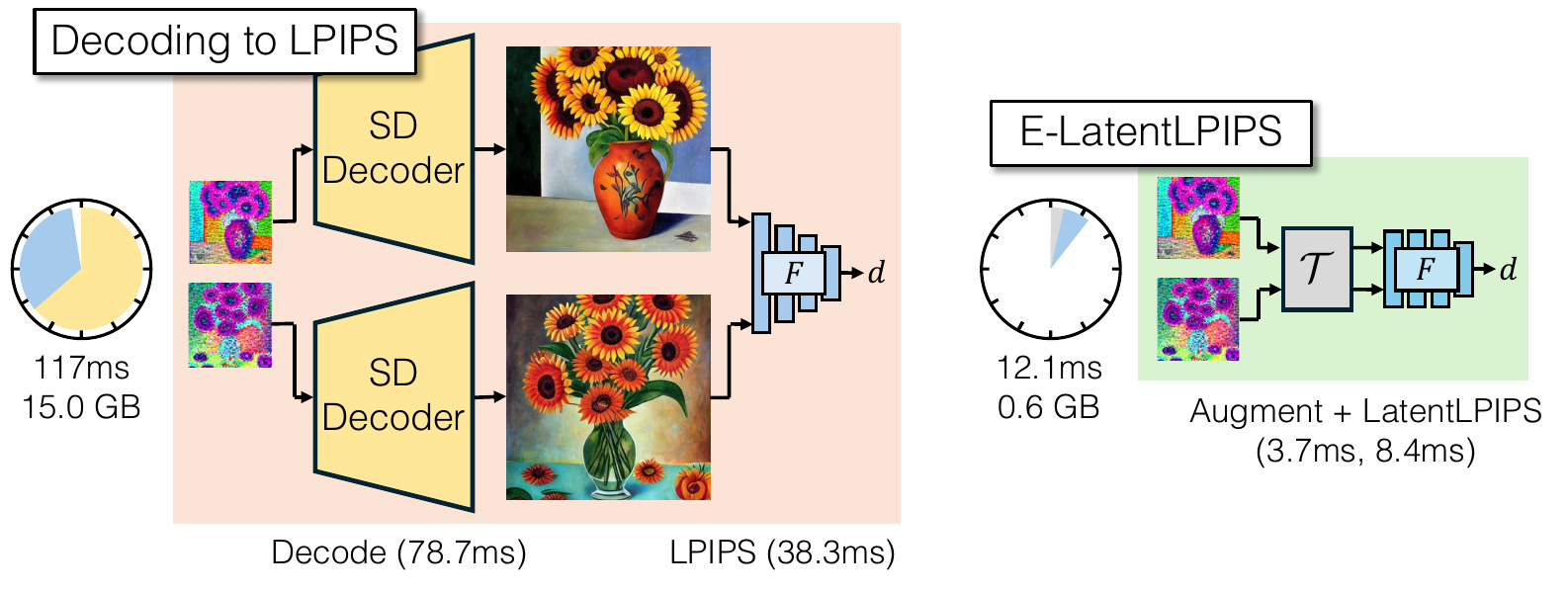}
    \vspace{-6pt}
    \caption{
    \textbf{E-LatentLPIPS for latent space distillation}. Training a single iteration with LPIPS~\cite{zhang2018unreasonable} takes 117ms and 15.0GB extra memory on NVIDIA A100, whereas our E-LatentLPIPS requires 12.1ms and 0.6GB on the same device. Consequently, E-latentLPIPS accelerates the perceptual loss computation time by $9.7\times$ compared to LPIPS, while simultaneously reducing memory consumption. 
    }
    \label{fig:LatentLPIPS}
    \vspace{-15pt}
\end{figure*}

\subsection{Ensembled-LatentLPIPS for Latent Space Distillation}
\label{sec:elatentlpips} The original LPIPS~\cite{zhang2018unreasonable} observes that the features from a pretrained classifier can be calibrated well enough to match human perceptual responses.
Moreover, LPIPS serves as an effective regression loss across many image translation applications~\cite{wang2018high,park2019semantic}. However, LPIPS, built to be used in the \textit{pixel} space, is unwieldy to use with a \textit{latent} diffusion model~\cite{rombach2022high}. As shown in Figure~\ref{fig:LatentLPIPS}, the latent codes must be decoded into the pixel space (e.g., $64\rightarrow 512$ resolution) before computing LPIPS with a feature extractor $F$ and a distance metric $\ell$.

\begin{equation}
\dist_\text{LPIPS}\big(\image_0, \image_1)=\ell\big(F(\text{Decode}^{8\times}(\image_0)), F(\text{Decode}^{8\times}(\image_1))\big)
\end{equation}

\noindent This defeats the primary motivator of LDMs, to operate in a more efficient latent space. As such, can we bypass the need to decode to pixels, and directly compute a perceptual distance in latent space?

\myparagraph{Learning LatentLPIPS.} We hypothesize that the same perceptual properties of LPIPS can hold for a function directly computed on latent space. Following the procedure from Zhang~\etal~\cite{zhang2018unreasonable}, we first train a VGG network~\cite{simonyan2014very} on ImageNet, but in the latent space of Stable Diffusion. We slightly modify the architecture by removing the 3 max-pooling layers, as the latent space is already $8\times$ downsampled, and change the input to 4 channels. We then linearly calibrate intermediate features using the BAPPS dataset~\cite{zhang2018unreasonable}. This successfully yields a function that operates in the latent space: $\dist_\text{LatentLPIPS}(\image_0, \image_1) = \ell(F(\image_0), F(\image_1))$.

Interestingly, we observe that while ImageNet classification accuracy in latent space is slightly lower on latent codes than on pixels, the perceptual agreement is retained. This indicates that while compression to latent space destroys some of the low-level information that helps with classification~\cite{ilyas2019adversarial}, it keeps the perceptually relevant details of the image, which we can readily exploit. Additional details are in the Appendix~\ref{appendix:elatentlpips}.

\begin{figure*}[t!]
    \centering
    \includegraphics[width=0.96\linewidth]{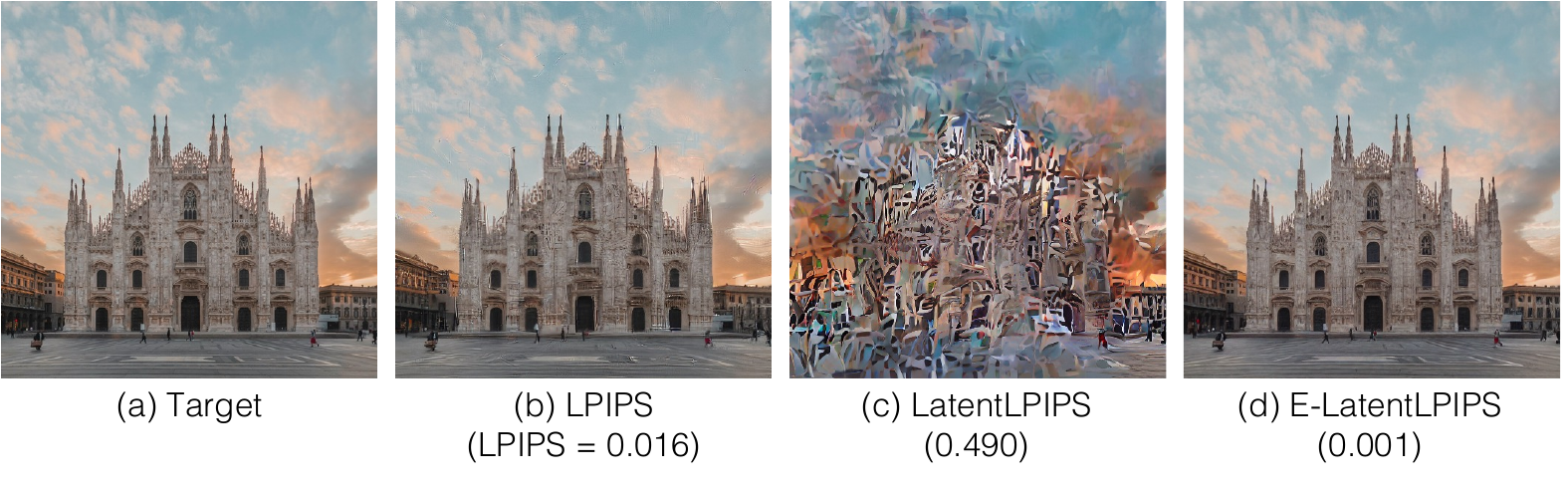}
    \vspace{-6pt}
    \caption{
    \textbf{Single image reconstruction}. To gain insight into the loss landscape of our regression loss, we conduct an image reconstruction experiment by directly optimizing a single latent with different loss functions.
    Reconstruction with LPIPS roughly reproduces the target image, but at the cost of needing to decode into pixels. LatentLPIPS alone cannot precisely reconstruct the image. However, our ensembled augmentation, E-LatentLPIPS, can more precisely reconstruct the target while operating directly in the latent space. 
    }
    \vspace{-4mm}
    \label{fig:singleimg}
\end{figure*}

\myparagraph{Ensembling.} We observe that the straightforward application of LatentLPIPS as the new loss function for distillation results in producing wavy, patchy artifacts. We further investigate this in a simple optimization setup, as shown in Figure~\ref{fig:singleimg}, by optimizing a randomly-sampled latent code towards a single target image. Here we aim to recover the target latent using different loss functions: $\arg \min_{\hat{\mathbf{x}}} \dist(\hat{\mathbf{x}}, \mathbf{x})$, where $\mathbf{x}$ is the target latent, $\hat{\mathbf{x}}$ the reconstructed latent, and $\dist$ either the original LPIPS or LatentLPIPS.
We observe that the single image reconstruction does not converge under LatentLPIPS (Figure~\ref{fig:singleimg}  (c)).
We hypothesize this limitation is due to a suboptimal loss landscape formed by the latent version of the VGG network.

Inspired by E-LPIPS~\cite{kettunen2019lpips}, we apply random differentiable augmentations~\cite{zhao2020differentiable, karras2020training}, general geometric transformations~\cite{karras2020training}, and cutout~\cite{devries2017improved}, to both generated and target latents. At each iteration, a random augmentation is applied to both generated and target latents. When applied to single image optimization, the ensemble strategy nearly perfectly reconstructs the target image, as shown in Figure~\ref{fig:singleimg} (d). The new loss is named Ensembled-LatentLPIPS, or E-LatentLPIPS for short.

\begin{equation}
\dist_\text{E-LatentLPIPS}(\image_0, \image_1) = \mathbb{E}_{\mathcal{T}} \Big[\ell\big(F(\mathcal{T}(\image_0)), F(\mathcal{T}(\image_1))\big)\Big],
\end{equation}

\noindent where $\mathcal{T}$ is a randomly sampled augmentation. Applying the loss function to ODE distillation:

\begin{equation}
   \mathcal{L}_\text{E-LatentLPIPS}\big(G,\noise, \prompt, \image \big) = \dist_\text{E-LatentLPIPS}(G(\noise, \prompt), \image),
  \label{eq:elatentlpips}
\end{equation}

\noindent where $\noise$ denotes a Gaussian noise, and $\image$ denotes its target latent. As illustrated in Figure~\ref{fig:LatentLPIPS} (right), compared to its LPIPS counterpart, the computation time is significantly lower, due to (1) not needing to decode to pixels (saving 79 ms for one image on an A100) and (2) (Latent)LPIPS itself operating at a lower-resolution latent code than in pixel space (38$\rightarrow$8 ms). While augmentation takes some time (4 ms), in total, perceptual loss computation is almost 10$\times$ cheaper (117$\rightarrow$ 12 ms) with our E-LatentLPIPS. In addition, memory consumption is greatly reduced (15$\rightarrow$0.6 GB).

Experimental results of Table~\ref{tab:ablation} demonstrate that learning the ODE mapping with E-LatentLPIPS leads to better convergence, exhibiting lower FID compared to other metrics such as MSE, Pseudo Huber loss~\cite{huber1992robust, song2023improved}, and the original LPIPS loss. For additional details regarding the toy reconstruction experiment and differentiable augmentations, please refer to the Appendix~\ref{appendix:elatentlpips}. %

\begin{figure*}[t!]
    \centering
    \includegraphics[width=\figwidth]{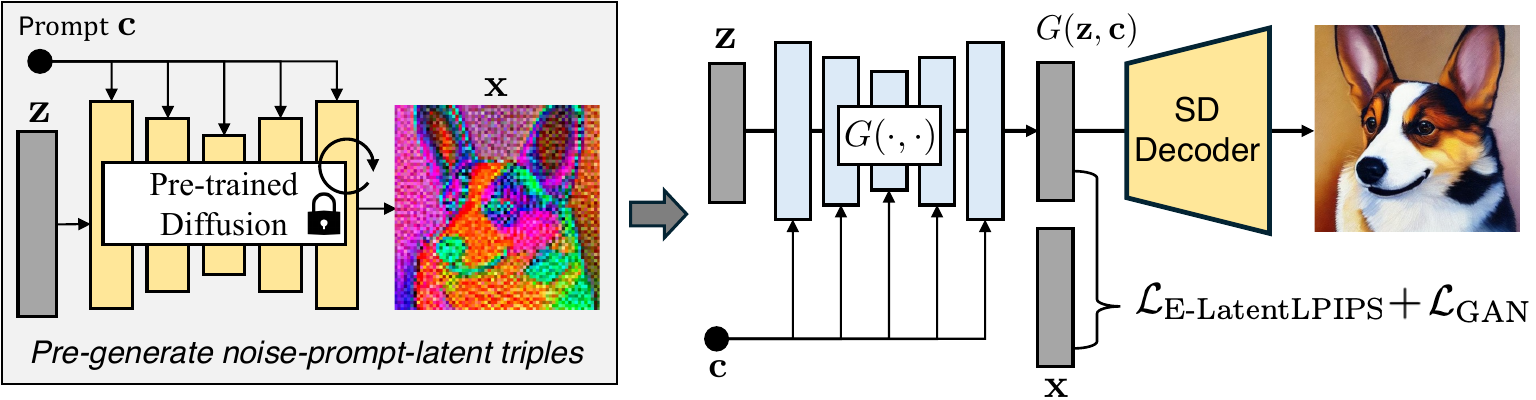}
    \vspace{-5pt}
    \label{fig:pipeline}
    \caption{
    {\bf Our \OursAcronym{} for one-step image synthesis.} First, we collect diffusion model output latents along with the input noises and prompts. Second, the generator is trained to map noise and prompt to the target latent using the E-LatentLPIPS regression loss~(Equation~\ref{eq:elatentlpips}) and the GAN loss~(Equation~\ref{eq:gan_loss}). While the output of the generator can be decoded by the SD latent decoder into RGB pixels, it is a compute intensive operation that is never performed during training. }
    \vspace{-15pt}
\end{figure*}

\subsection{Conditional Diffusion Discriminator}
\label{sec:discriminator}
In Sections~\ref{sec:generator} and~\ref{sec:elatentlpips}, we have elucidated that diffusion distillation can be achieved by framing it as a paired noise-to-latent translation task. Motivated by the effectiveness of conditional GAN for paired image-to-image translation~\cite{isola2017image}, 
we employ a conditional discriminator. The conditions for this discriminator include not only the text description~$\prompt$ but also the Gaussian noise $\noise$ provided to the generator. Our new discriminator incorporates the aforementioned conditioning while leveraging the pre-trained diffusion weights. Formally, we optimize the following minimax objective for the generator $G$ and discriminator $D$: 
\begin{equation}
    \min_{G}\max_{D}~\mathbb{E}_{\prompt, \noise, \image }[\log(D( \prompt, \noise, \image))] + \mathbb{E}_{ \prompt, \noise}[\log(1 - D(\prompt, \noise, G(\noise, \prompt)))].
    \label{eq:gan_minimax}
\end{equation}
For the generator, we minimize the following non-saturating GAN loss~\cite{goodfellow2016nips}.
\begin{equation}
    \mathcal{L}_\text{GAN}(G, \noise, \prompt, \image   ) =  - \mathbb{E}_{ \prompt, \noise} \big[\log(D(\prompt, \noise, G(\noise, \prompt))) \big].
    \label{eq:gan_loss}
\end{equation}
The final loss for the generator is $\mathcal{L}_G = \mathcal{L}_\text{E-LatentLPIPS} + \lambda_{\text{GAN}} \mathcal{L}_\text{GAN}$. 
We provide more details on the discriminator and loss functions. 

\begin{figure*}[t!]
    \centering
    \includegraphics[width=\figwidth]{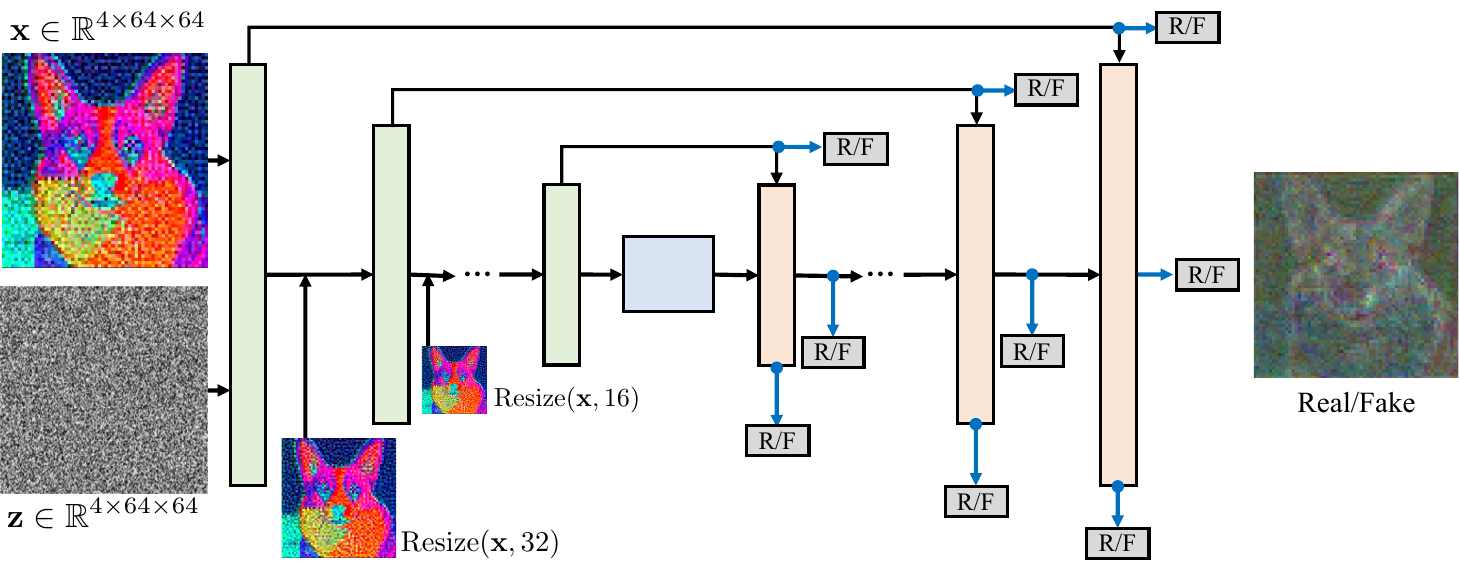}
    \vspace{-5pt}
    \caption{
    {\bf Our multi-scale conditional discriminator design.} We reuse the pretrained weights from the teacher model's U-Net and augment it with multi-scale input and output branches. Concretely, we feed the resized version of input latents to each downsampling block of the encoder. For the decoder part, we enforce the discriminator to make real/fake predictions at three places at each scale: before, at, and after the skip connection. %
    This multi-scale adversarial training further improves image quality. 
    }%
        \label{fig:discriminator}
\vspace{-15pt}
\end{figure*}

\myparagraph{\textbf{Initialization from a pre-trained diffusion model.}} We demonstrate that initializing the discriminator weights with a pre-trained diffusion model is effective for diffusion distillation. Compared to the implementation of GigaGAN discriminator~\cite{kang2023scaling}, using a pre-trained Stable Diffusion 1.5 U-Net~\cite{stablediffusion1.5} and finetuning the model as the discriminator in the latent space results in superior FID in Table~\ref{tab:discriminator_ablation}. The adversarial loss is computed independently at each location of the U-Net discriminator output. Note that the original U-Net architecture conditions on text but not on the input noise map $\noise$. We further modify the discriminator architecture to support $\noise$ conditioning, simply by adding the input with $\noise$ processed through a single convolution layer with zero initialization in the channel dimension. Note that the text conditioning for the diffusion discriminator is naturally carried out by the built-in cross-attention layers in the Stable Diffusion U-Net. We observe moderate improvement across all metrics.

\myparagraph{Single-sample R1 regularization.} 
While the conditional U-Net discriminator from pre-trained diffusion weights already achieves competitive results on the zero-shot COCO2014~\cite{lin2014microsoft} benchmark, we have noticed considerable training variance across different runs, likely due to the absence of regularization and unbounded gradients from the discriminator. To mitigate this, we introduce R1 regularization~\cite{Mescheder2018ICML} on each mini-batch for training the diffusion discriminator. However, introducing R1 regularization increases GPU memory consumption, posing a practical challenge, especially when the discriminator is a high-capacity U-Net. To minimize memory consumption and accelerate training, we not only adopt lazy regularization~\cite{karras2020analyzing} with an interval of 16, but also apply R1 regularization only to a single sample of each mini-batch. In addition to improved stability, we also observe that the single-sample R1 regularization results in better convergence, as shown in Table~\ref{tab:discriminator_ablation}. 

\myparagraph{Multi-scale in-and-out U-Net discriminator.} GigaGAN~\cite{kang2023scaling} observes that the GAN discriminator tends to focus on a particular frequency band, often overlooking high-level structures, and introduces a multi-scale discriminator to address this issue. Similarly, we propose a new U-Net discriminator design,   as shown in Figure~\ref{fig:discriminator}, which enforces independent real/fake prediction at various segments of the U-Net. Specifically, we modify the U-Net encoder to receive resized inputs at each downsampling layer and attach three readout layers at each scale of the U-Net decoder to make independent real/fake predictions, from the U-Net skip connection features, the upsampled features from the U-Net bottleneck, and the combined features. At a high level, the new design enforces that all U-Net layers participate in the final prediction, ranging from shallow skip connections to deep middle blocks. This design enhances low-frequency structural consistency and significantly increases FIDs, as observed in Table~\ref{tab:discriminator_ablation}. 

\myparagraph{Mix-and-match augmentation.} 
To further encourage the discriminator to focus on text alignment and noise conditioning, we introduce mix-and-match augmentation for discriminator training, similar to GigaGAN~\cite{kang2023scaling} and earlier text-to-image GAN works~\cite{reed2016generative,zhang2017stackgan}. During discriminator training, we replace a portion of the generated latents with random, unrelated latents from the target dataset while maintaining the other conditions unchanged. This categorizes the replaced latents as fake, since the alignments between the latent and its paired noise and text are incorrect, thereby fostering improved alignments. Additionally, we make substitutions to text and noise, contributing to the overall enhancement of the conditional diffusion discriminator.

\section{Experiments}
Here, we first study the effectiveness of our algorithmic designs with a systematic ablation study in Section~\ref{sec:ablation}. Next, We compare our method with leading one-step generators using a standard benchmark regarding image quality, text alignment, and inference speed in Section~\ref{sec:comparison}. We then present human preference evaluation results in Section~\ref{sec:human_evaluation}. Additionally, we provide visual comparisons (Section~\ref{sec:visual_analysis}) and report the training speed (Section~\ref{sec:training_speed}).

\myparagraph{Training details.}
We distill Stable Diffusion 1.5 into our one-step generator and train the model on two ODE datasets with different classifier-free guidance (CFG), namely, the SD-CFG-3 dataset with 3 million noise-latent pairs and the SD-CFG-8 dataset with 12 million pairs. We use the prompts from the LAION-aesthetic-6.25 and -6.0 datasets to create the SD-CFG-3 and SD-CFG-8 datasets, respectively, and simulate the ODE using 50 steps of DDIM~\cite{song2021denoising}. To demonstrate the effectiveness of \OursAcronym{} for a larger text-to-image model, we distill SDXL-Base-1.0~\cite{podell2024sdxl} into Diffusion2GAN using 8 million noise-latent pairs named SDXL-CFG-7 dataset. These pairs were generated by SDXL-Base-1.0 using prompts from the LAION-aesthetic-6.0 dataset. We simulate the ODE of SDXL-Base-1.0 using 50 steps of DDIM. For further details on hyperparameters, please refer to the Appendix~\ref{appendix:train_eval_details}. Notably, we did not use any real images from the LAION dataset.

\myparagraph{Evaluation protocols.}
We evaluate our model on two widely used datasets, COCO2014 and COCO2017. We include the evaluation on COCO2017, as progressive distillation~\cite{salimans2022progressive} and DPM solver~\cite{lu2022dpm} only report results on this dataset. We use FID~\cite{Heusel2017GANsTB} and CLIP-score~\cite{hessel2021clipscore} to assess image realism and text-to-image alignment. Following GigaGAN's protocol~\cite{kang2023scaling}, we resize the generated images 512px to 256px, reprocess them to 299px, and then feed them into the InceptionV3 network for FID and Precision $\&$ Recall calculations~\cite{sajjadi2018assessing,Kynknniemi2019ImprovedPA}. FID~\cite{Heusel2017GANsTB,parmar2021cleanfid} is computed on 40,504 real images from the COCO2014 validation dataset and 30,000 fake images generated using 30,000 randomly sampled COCO2014 validation prompts, while Precision $\&$ Recall are calculated on 10,000 images due to their heavy computation. For COCO2017 dataset, we use 5,000 image-text pairs for FID and CLIP-score calculations. Precision $\&$ Recall metrics on COCO2017 are omitted, as we heuristically find that 5,000 real samples are insufficient to yield valid measurements of image fidelity and diversity. Instead, we introduce a new diversity score, calculated using DreamSim~\cite{fu2023dreamsim}, to quantify the range of variation in the generated images. Note that the resizing processes in the evaluation pipeline are performed using an antialiasing bicubic resizer, as recommended by Parmar~\etal~\cite{parmar2021cleanfid}.

\begin{table}[t]
\centering
\caption{\textbf{Ablation study on SD-CFG-3 dataset.} We distill Stable Diffusion 1.5~\cite{stablediffusion1.5} into one-step generators using ODE distillation loss~(Equation~\ref{eq:our_regression_loss}). All models are trained with a batch size of 256 for 20k iterations using 8 A100-80GB GPUs, except for the LPIPS model and the larger batch-size model. For the LPIPS model, we employ a batch size of 64 and accumulate the gradients four times. This adjustment is necessary due to the LPIPS model consuming 62GB of GPU memory per A100-80GB when applied to a batch size of 64, whereas the other models require nearly 68GB of per GPU memory for 256 batch training. Our E-LatentLPIPS achieves stronger performance than traditional LPIPS without the need to decode to pixels.}
\vspace{-2mm}
\resizebox{1.0\textwidth}{!}{
\begin{tabular}{lcccccccc}
\cmidrule[1.0pt]{1-8}
Method (Loss function) & Loss space & ~img/sec$~(\uparrow)$~ & Batch size ~ & FID$~(\downarrow)$~ & ~CLIP$~(\uparrow)$~ & ~Pre.$~(\uparrow)$~ & ~Rec.$~(\uparrow)$~\\
\cmidrule[0.5pt]{1-8}
ODE distillation (LPIPS~\cite{zhang2018unreasonable}) & Pixel & 40.0 & 256 & 25.94 & 0.288 & 0.60 & 0.53 & \\
ODE distillation (MSE) & Latent & 138.4 & 256 & 110.55 & 0.222 & 0.21 & 0.33 \\
ODE distillation (Pseudo Huber~\cite{song2023improved}) & Latent & 144.2 & 256 & 87.60 & 0.230 & 0.29 & 0.40 \\
ODE distillation (LatentLPIPS) & Latent & 139.9 & 256 & 67.17 & 0.244 & 0.46 & 0.54 \\
ODE distillation (E-LatentLPIPS) & Latent & 127.5 & 256 & \textbf{22.95} & \textbf{0.299} & \textbf{0.62} & \textbf{0.58} \\ %
\cmidrule[0.5pt]{1-8}
\hspace{2mm} $\rightarrow$ larger batch-size (8$\times$ more GPUs) & Latent & 128.0 & 2048 & \textbf{14.72} & 0.292 & \textbf{0.66} & \textbf{0.65} \\
\cmidrule[1.0pt]{1-8}
\end{tabular}
}
\vspace{-1mm}
\label{tab:ablation}
\end{table}

\subsection{Effectiveness of Each Component}
\label{sec:ablation}
In Table~\ref{tab:ablation}, we conduct an ablation study on the choice of distance metric for ODE distillation training. We consider L1, Pseudo Huber~\cite{song2023improved}, LPIPS, LatentLPIPS, and our E-LatentLPIPS metrics. As shown in Table~\ref{tab:ablation}, ODE distillation using MSE~\cite{luhman2021knowledge} %
achieves worse results on large-scale text-to-image datasets. Also, introducing the Pseudo Huber metric improves FID significantly~\cite{song2023consistency},  
but it remains insufficient. However, if we apply a perceptual loss, such as pixel space LPIPS and latent space E-LatentLPIPS, the ODE distillation presents FID near 20$\sim$25, even trained using a small batch size. This suggests that the noise-to-image translation task holds promise, and it would give better results once we introduce a conditional discriminator to further improve the image quality.

Table~\ref{tab:discriminator_ablation} presents the ablation study regarding each component of Diffusion2GAN's discriminator. All generators are initialized with the pre-trained weights of the best performing ODE distilled generator shown in Table~\ref{tab:ablation}. We compare our diffusion-based discriminator to the state-of-the-art GigaGAN discriminator~\cite{kang2023scaling}. As shown in Table~\ref{tab:discriminator_ablation}, each component of Diffusion2GAN plays a crucial role in enhancing FID and CLIP-score. Diffusion2GAN surpasses ODE distillation with the GigaGAN discriminator while narrowing the performance gap with the Stable Diffusion 1.5.

\begin{table}[t]
\centering
\caption{\textbf{Ablation study on SD-CFG-3 dataset.} All models are initialized with the weights of a pre-trained ODE distillation model targeting Stable Diffusion 1.5~\cite{stablediffusion1.5} and trained with a batch size of 256 using 16 A100-80GB GPUs. Each proposed component plays a crucial role in improving both FID~\cite{Heusel2017GANsTB} and CLIP-score~\cite{hessel2021clipscore}.}
\vspace{-2mm}
\resizebox{0.97\textwidth}{!}{
\begin{tabular}{lcccc}
\cmidrule[1.0pt]{1-5}
Method & FID-30k$~(\downarrow)$~ & ~CLIP-30k$~(\uparrow)$~ & ~Precision$~(\uparrow)$~ & ~Recall$~(\uparrow)$~ \\
\cmidrule[0.5pt]{1-5}
ODE distillation (E-LatentLPIPS) &  14.72 &  0.292 & 0.66 & 0.65 \\
~+~GigaGAN D~\cite{kang2023scaling} & 13.97 & 0.293 & 0.68 & 0.64 \\ 
\cmidrule[1.0pt]{1-5}
ODE distillation (E-LatentLPIPS) &  14.72 &  0.292 & 0.66 & 0.65 \\
~+~Diffusion D & 12.04 & 0.300 & 0.70 & 0.65 \\
~+~$\textbf{z}$ conditional D & 11.97 & 0.302 & 0.70 & 0.65 \\
~+~Single-sample R1 & 10.60 & 0.303 & 0.73 & 0.65 \\
~+~Multi-scale training & 9.58 & 0.308 & 0.72 & {\bf 0.66} \\
~+~Mix-and-match augmentation & {\bf 9.45} & {\bf 0.310} & {\bf 0.73} & 0.65\\
\cmidrule[1.0pt]{1-5}
Stable Diffusion 1.5~\cite{stablediffusion1.5} (Teacher)  & 8.74 & 0.312 & 0.72 & 0.67\\
\cmidrule[1.0pt]{1-5}
\end{tabular}
}
\vspace{-2mm}
\label{tab:discriminator_ablation}
\end{table}

\subsection{Comparison with Distilled Diffusion Models}
\label{sec:comparison}
\myparagraph{Distilling Stable Diffusion 1.5.} We compare Diffusion2GAN with cutting-edge diffusion distillation models on both COCO2014 and COCO2017 benchmarks in Tables~\ref{tab:coco2014_benchmark} and~\ref{tab:coco5k_benchmark}, respectively. While InstaFlow-0.9B can attain an FID of 13.10 on COCO2014 and 23.4 on COCO2017, Diffusion2GAN can more efficiently deal with the ODE distillation problem, achieving an FID of 9.29 and 19.5, respectively. Similar to our method, UFOGen~\cite{xu2023ufogen}, MD-UFOGen~\cite{zhao2023mobilediffusion}, DMD~\cite{yin2023onestep}, and ADD-M~\cite{sauer2023adversarial} use extra diffusion models for adversarial training or distribution matching. Although these models achieve lower FIDs compared to InstaFlow-0.9B, Diffusion2GAN still outperforms them, as Diffusion2GAN is trained to closely follow the original trajectory of the teacher diffusion model, thus mitigating the diversity collapse issue while maintaining high visual quality. Note that the concurrent work ADD-M exhibits a higher CLIP-score compared to \OursAcronym{}. We hypothesize this is because ADD-M conditions the discriminator using both image and text embeddings, as shown in Table 1(b) of the ADD-M paper~\cite{sauer2023adversarial}. While \OursAcronym{} focuses on efficiency and prefers not to produce the pixels required for obtaining image embeddings, in the SDXL distillation experiment, we show that \OursAcronym{} demonstrates better image quality and text-to-image alignment compared to ADD-XL, also known as SDXL-Turbo.

\begin{table}[hp!]
\centering
\caption{{\bf Comparison to recent text-to-image models on COCO2014}. We distill Stable Diffusion 1.5~\cite{stablediffusion1.5} into Diffusion2GAN on the SD-CFG-3 dataset with a batch size of 1024 using 64 A100-80GB GPUs. \OursAcronym{} significantly outperforms the leading one-step diffusion distillation generators.}
\vspace{-2mm}
\resizebox{0.88\textwidth}{!}{
\begin{tabular}{lccccc}
\cmidrule[1.0pt]{1-5}
 Multi-step generator & ~Type~ &  ~\# Param.~ & ~FID-30k~($\downarrow$)~ & ~Inference time (s)~\\
\cmidrule[1.0pt]{1-5}
DALL$\cdot$E 2~\cite{ramesh2022hierarchical} & Diffusion & 5.5B & 10.39 & -  \\
Imagen~\cite{saharia2022photorealistic} & Diffusion & 3.0B & 7.27 & 9.1  \\
Stable Diffusion 1.5~\cite{stablediffusion} & Diffusion & 0.9B & 8.74 & 2.59 \\
PIXART-$\alpha$~\cite{chen2023pixart} & Diffusion & 0.6B & 10.65 & - \\
\cmidrule[1.0pt]{1-5}
 One-step generator & ~Type~ &  ~\# Param.~ & ~FID-30k~($\downarrow$)~ & ~Inference time (s)~\\
\cmidrule[1.0pt]{1-5}
GigaGAN~\cite{kang2023scaling} & GAN & 1.0B & \textbf{9.09} & 0.13\\
StyleGAN-T~\cite{Sauer2023StyleGANTUT} & GAN & 1.0B & 13.90 & 0.10\\
InstaFlow-0.9B~\cite{liu2023insta} & Distillation & 0.9B & 13.10 & 0.09\\
UFOGen~\cite{xu2023ufogen} & Distillation & 0.9B & 12.78 & 0.09\\
MD-UFOGen~\cite{zhao2023mobilediffusion} & Distillation & 0.4B & 11.67 & \textbf{< 0.09}\\
DMD~\cite{yin2023onestep} & Distillation & 0.9B & 11.49 & 0.09\\
\textbf{\OursAcronym{}} & Distillation & 0.9B & 9.29 & 0.09\\
\cmidrule[1.0pt]{1-5}
\end{tabular}}
\vspace{-2mm}
\label{tab:coco2014_benchmark}
\end{table}

\begin{table}[hp!]
\centering
\caption{{\bf Comparison to recent text-to-image models on COCO2017}. On the SD-CFG-3, Diffusion2GAN, distilled from Stable Diffusion 1.5~\cite{stablediffusion1.5}, demonstrates better performance over UFOGen~\cite{xu2023ufogen}. While Diffusion2GAN presents slightly better FID~\cite{Heusel2017GANsTB} than ADD-M~\cite{sauer2023adversarial}, it exhibits a lower CLIP-score~\cite{hessel2021clipscore}.}
\vspace{-2mm}
\resizebox{0.96\textwidth}{!}{
\begin{tabular}{lccccc}
\cmidrule[1.0pt]{1-5}
 Model & ~\# Step~ & ~FID-5k~($\downarrow$)~ & ~CLIP-5k~($\uparrow$)~ & ~Inference time (s)~\\
\cmidrule[1.0pt]{1-5}
DPM solver~\cite{lu2022dpm} & 25 &  20.1 & 0.318 & 0.88 \\
Progressive distillation~\cite{meng2023distillation} & 4 &  26.4 & 0.300 & 0.21\\
CFG-Aware distillation~\cite{li2024snapfusion} & 8 & 24.2 & 0.300 & 0.34\\
InstaFlow-0.9B~\cite{liu2023insta} & 1 & 23.4 &  0.304 & \textbf{0.09} \\
UFOGen~\cite{xu2023ufogen} & 1 & 22.5 & 0.311 & \textbf{0.09} \\
ADD-M~\cite{sauer2023adversarial} & 1 & 19.7 &  \textbf{0.326} & \textbf{0.09}\\
\textbf{\OursAcronym{}} & 1 & \textbf{19.5} & 0.311 & \textbf{0.09} \\
\cmidrule[0.5pt]{1-5}
Stable Diffusion 1.5~\cite{stablediffusion1.5} (Teacher) & 50 & 19.1 & 0.313 & 2.59 \\ 
\cmidrule[1.0pt]{1-5}
\end{tabular}}
\vspace{-2mm}
\label{tab:coco5k_benchmark}
\end{table}

\begin{table}[hp!]
\centering
\caption{{\bf Comparison to recent text-to-image models on COCO2017}. On the SDXL-CFG-7 dataset, Diffusion2GAN, distilled from SDXL-Base-1.0~\cite{podell2024sdxl}, demonstrates better FID and CLIP-score~\cite{hessel2021clipscore} over SDXL-Turbo~\cite{sauer2023adversarial} and SDXL-Lightning~\cite{lin2024sdxl}. Our proposed diversity score, DreamDiv, confirms that SDXL-Diffusion2GAN generates more diverse images compared to SDXL-Turbo while exhibiting better text-to-image alignment compared to both SDXL-Turbo and SDXL-Lightning.}
\vspace{-2mm}
\resizebox{1.0\textwidth}{!}{
\begin{tabular}{lccccc}
\cmidrule[1.0pt]{1-6}
 Model &  ~\# Step~  & FID-5k$~(\downarrow)$~ & ~CLIP-5k$~(\uparrow)$~ & ~DreamDiv-5k$~(\uparrow)$~ & ~DreamSim-5k$~(\downarrow)$~\\
\cmidrule[1.0pt]{1-6}
SDXL-Turbo~\cite{sauer2023adversarial} & 1 & 28.10 & 0.342 & 0.232 & 0.368\\
SDXL-Lightning~\cite{lin2024sdxl} & 1 & 30.14 & 0.324 & \textbf{0.315} & 0.345\\
\textbf{SDXL-\OursAcronym{}} & 1 &\textbf{25.49} & \textbf{0.347} & 0.268 & \textbf{0.284}\\
\cmidrule[0.5pt]{1-6}
SDXL-Base-1.0~(Teacher)~\cite{podell2024sdxl} & 50 & 25.56  & 0.346 & 0.338 & 0.0 \\
\cmidrule[1.0pt]{1-6}
\end{tabular}}
\vspace{-2mm}
\label{tab:sdxl_benchmark}
\end{table}

\myparagraph{Distilling SDXL-Base-1.0.} To demonstrate the effectiveness of Diffusion2GAN for a larger text-to-image model, we distill SDXL-Base-1.0~\cite{podell2024sdxl} into Diffusion2GAN and evaluate its performance using FID and CLIP-score on COCO2017. Through empirical analysis, we have observed that Recall~\cite{Kynknniemi2019ImprovedPA} is inadequate for measuring image diversity with only 5,000 real images. As an alternative to Recall, we generate 8 images per prompt and calculate average pairwise perceptual distance, measured by DreamSim~\cite{fu2023dreamsim}, to quantify the diversity of the generated images. We name this metric \textit{DreamDiv}. The rationale behind this metric is that diversity can be captured by perceptual dissimilarity within the same prompt. A similar LPIPS-based diversity metric has been widely used in multimodal image-to-image translation~\cite{zhu2017toward,huang2018multimodal}. Table~\ref{tab:sdxl_benchmark} presents that SDXL-Diffusion2GAN achieves comparable FID and CLIP-score with the teacher SDXL-Base-1.0, while exhibiting higher DreamDiv compared to SDXL-Turbo. SDXL-Lightning shows higher DreamDiv than SDXL-Diffusion2GAN but a lower CLIP-score compared to SDXL-Diffusion2GAN, indicating that the observed high diversity of SDXL-Lightning is due to poor text-to-image alignment. We recommend that DreamDiv should be reported with the CLIP-score to prevent a scenario where DreamDiv is high due to poor text-to-image alignment. 

To quantify the capability of learning the diffusion teacher's ODE trajectory, we introduce \textit{DreamSim-5k}. Specifically, we simulate the ODE of both a target diffusion model and each one-step generator using 5k randomly sampled noises and COCO2017 prompts. DreamSim-5k score is then computed by averaging DreamSim~\cite{fu2023dreamsim} between pairs of images generated from the same noise. A lower DreamSim-5k indicates better preservation of the noise-image mapping of the teacher diffusion model. Compared to SDXL-Turbo and SDXL-Lightning, SDXL-Diffusion2GAN demonstrates better ability to learn the noise-image mapping of teacher SDXL-Base-1.0 as shown in Table~\ref{tab:sdxl_benchmark}.

\subsection{Human Preference Evaluation}
\label{sec:human_evaluation}

\begin{figure*}[!ht]
    \centering
    \includegraphics[width=\figwidth]{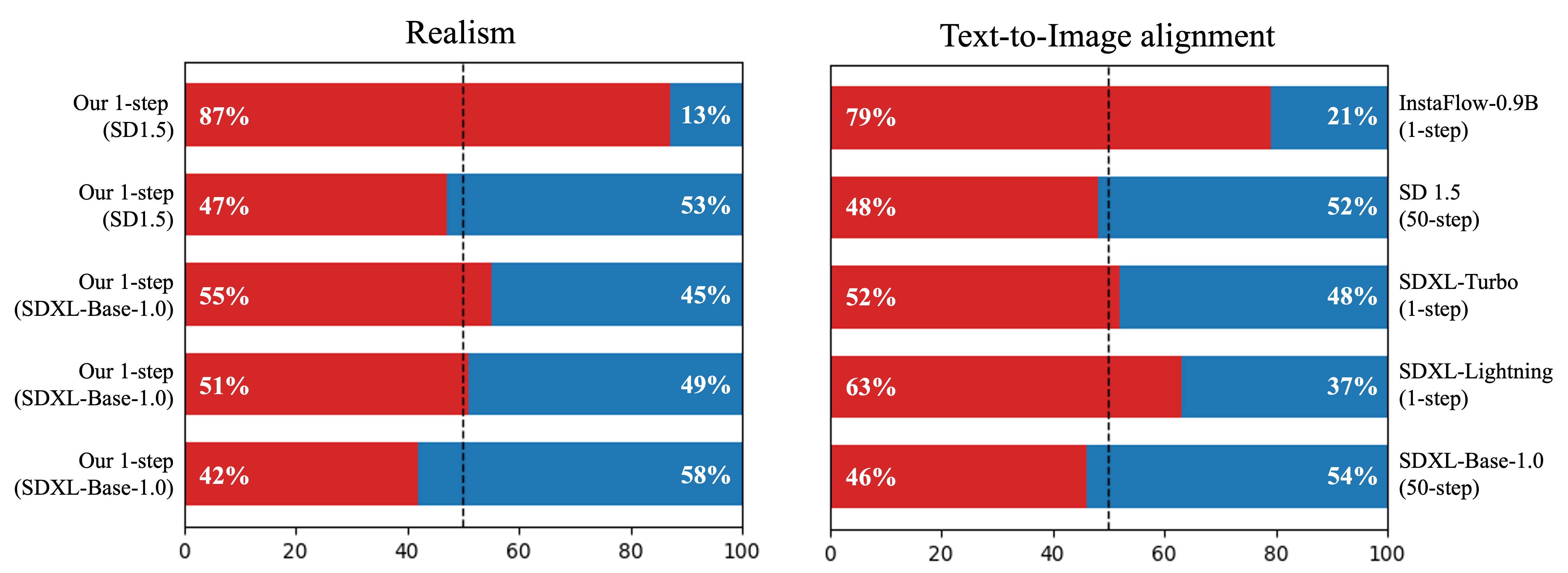}
    \caption{
    {\bf Human preference evaluation.} We evaluate human preferences using images generated from 128 PartiPrompts~\cite{yu2022scaling}. Annotators evaluate both the image quality and text-to-image alignment. All models, except for Stable Diffusion 1.5 (SD 1.5)~\cite{stablediffusion1.5} and SDXL-Base-1.0~\cite{podell2024sdxl}, generate images using a single forward evaluation. SD 1.5 and SDXL-Base-1.0 generated images using 50 steps of DDIM~\cite{song2021denoising}.
    }
    \vspace{-2mm}
    \label{fig:human_eval}
\end{figure*}

We conduct human preference evaluations following the procedure described in the LADD paper~\cite{sauer2024fast}, using 128 PartiPrompts~\cite{yu2022scaling} to assess preferences for image realism and text-to-image alignment. We compare \OursAcronym{} with its baselines and teacher model as shown in Figure~\ref{fig:human_eval} and ensure a fair comparison by comparing models distilled from the same teacher model.

For Stable Diffusion 1.5 distillation, \OursAcronym{} shows better human preferences for both image realism and text-to-image alignment over InstaFlow-0.9B. For SDXL-Base-1.0 distillation, SDXL-\OursAcronym{} demonstrates comparable or superior image realism and text-to-image alignment compared to SDXL-Turbo and SDXL-Lightning. Notably, one aspect (realism or text-to-image alignment) of SDXL-Diffusion2GAN tends to be preferred over SDXL-Turbo and SDXL-Lightning when the other aspect is comparable to the baselines. While our proposed one-step generators exhibit better realism and text-to-image alignment than previous one-step models, the multi-step teacher models are still preferred overall. We leave the future improvement of Diffusion2GAN for future work.

\subsection{Visual Analysis}
\label{sec:visual_analysis}
\myparagraph{Distilling Stable Diffusion 1.5.} We visually compare our model with Stable Diffusion 1.5~\cite{stablediffusion1.5}, LCM-LoRA~\cite{luo2023latentlora}, and InstaFlow~\cite{liu2023insta} in Figure~\ref{fig:visual_comparison}.  As diffusion models tend to generate more photo-realistic images with an increased scale of classifier-free guidance (CFG)~\cite{ho2022classifier}, we train our \OursAcronym{} using the SD-CFG-8 dataset and compare it against Stable Diffusion 1.5 using the same guidance scale of 8. For LCM-LoRA and InstaFlow, we follow their best settings to ensure a fair comparison. Our findings indicate that our model produces images with enhanced realism compared to the other distillation baselines, while maintaining the overall layout of the target images generated by the Stable Diffusion teacher. We could not compare Diffusion2GAN with more advanced distillation models, as pre-trained weights were not publicly available. Therefore, we compare these models only quantitatively in Tables~\ref{tab:coco2014_benchmark} and \ref{tab:coco5k_benchmark}.

\myparagraph{Distilling SDXL-Base-1.0.} We compare Diffusion2GAN, distilled on SDXL-Base-1.0, with two concurrent works, SDXL-Turbo~\cite{sauer2023adversarial} and SDXL-Lightning~\cite{lin2024sdxl}, in Figure~\ref{fig:visual_comparison2}. While the images produced by all these models generally appear realistic, SDXL-Diffusion2GAN is the only model capable of following the ODE trajectory of the teacher diffusion model. SDXL-Diffusion2GAN tends to generate more diverse images than SDXL-Turbo while exhibiting more plausible structural features than SDXL-Lightning.

\begin{table}[tp!]
\vspace{-1mm}
\begin{minipage}[b]{0.52\textwidth}
\centering
    \resizebox{1.0\textwidth}{!}{
    \begin{tabular}{lccc}
      \cmidrule[1.0pt]{1-4}
      \multirow{2}*[0.0ex]{Model}~&~Total~&~$\#$~saved~&~\multirow{2}*[0.0ex]{FID-50k~($\downarrow$)}~\\
      &~$\#$~NFE~&~images~& \\
      \cmidrule[1.0pt]{1-4}
      Consist. Distill.~\cite{song2023consistency} & 819.2M & 50k & 3.67\\
      \cmidrule[0.5pt]{1-4}
      \multirow{2}*[-2.5ex]{ODE distillation} & 78.6M &  50k & 8.51 \\
         & 80.3M &  100k & 5.62 \\
       & 83.8M &  200k & 3.85 \\
       \multirow{2}*[3.0ex]{(LPIPS~\cite{zhang2018unreasonable})}& 94.3M &  500k & 3.25 \\
       & 111.8M &  1.0M & \textbf{3.16} \\
      \cmidrule[1.0pt]{1-4}
    \end{tabular}}
    \caption{LPIPS regression achieves better FID~\cite{Heusel2017GANsTB} than Consistency Distillation~\cite{song2023consistency} on CIFAR10~\cite{Krizhevsky2009LearningML}, while requiring fewer number of function evaluations (NFE) for both ODE pair generation and model training.}
    \label{table:cifar10_exp}
\end{minipage}
\hfill
\begin{minipage}[b]{0.46\textwidth}
\centering
    \resizebox{1.0\textwidth}{!}{
    \begin{tabular}{lcc}
      \cmidrule[1.0pt]{1-3}
      {Model}~&~A100 days~&~FID-30k~($\downarrow$)~\\
      \cmidrule[1.0pt]{1-3}
       InstaFlow-0.9B~\cite{liu2023insta} & 183.2 &  13.10 \\
       ODE distillation & 36.0 &  15.94 \\
       \textbf{Diffusion2GAN} & 43.6 & \textbf{9.29} \\
      \cmidrule[1.0pt]{1-3}
    \end{tabular}}
    \caption{Diffusion2GAN requires fewer A100 GPU days for training and attains a significantly lower FID compared to InstaFlow~\cite{liu2023insta}. The number of A100 days and FID for InstaFlow are obtained from the original paper. We train the ODE distillation model and \OursAcronym{} using a batch size of 256 for 150k and 160k iterations, respectively.}
    \label{tab:instaflow_exp}
\end{minipage}
\end{table}

\subsection{Training Speed}
\label{sec:training_speed}
Even including the cost of preparing the ODE dataset, Diffusion2GAN converges more efficiently than the existing distillation methods. On the CIFAR10 dataset, we compare the total number of function evaluations of the generator network in total training. We find that training with the LPIPS loss on 500k teacher outputs already surpasses the FID of Consistency Distillation~\cite{song2023consistency} at a fraction of the total compute budget (Table~\ref{table:cifar10_exp}). On text-to-image synthesis, our full version Diffusion2GAN achieves superior FID compared to InstaFlow, all while utilizing considerably fewer GPU days~(Table~\ref{tab:instaflow_exp}).
\vspace{-1.5mm}

\section{Discussion and Limitations}
\label{sec:discussion}
We have proposed a new framework \textit{\OursAcronym{}} for distilling a pre-trained multi-step diffusion model into a one-step generator trained with conditional GAN and perceptual losses. 
Our study shows that separating generative modeling into two tasks—first identifying correspondences and then learning a mapping—allows us to use different generative models to improve the performance-runtime tradeoff. Our one-step model is not only beneficial for interactive image generation but also offers the potential for efficient video and 3D applications.

\myparagraph{\textbf{Limitations.}}
Although our method achieves faster inference while maintaining image quality, it does have several limitations. First, our current approach simulates a fixed classifier-free guidance scale, a common technique for adjusting text adherence, but does not support varying CFG values at inference time. Exploring methods like guided distillation~\cite{meng2023distillation} could be a promising direction. Second, as our method distills a teacher model, the performance limit of our model is bound by the quality of the original teacher's output. To enhance the quality of generated noise-image pairs, employing advanced diffusion models like EDM2~\cite{Karras2024edm2}, which is better compatible with deterministic sampling, could be advantageous. Additionally, leveraging real text and image pairs is a potential avenue to learn a student model that outperforms the original teacher model. Third, Diffusion2GAN only supports one-step image synthesis as it was trained to translate given noise into an RGB image directly. However, extending Diffusion2GAN to multi-step generation could result in future performance improvement. Last, while \OursAcronym{} alleviates the diversity drop by introducing ODE distillation loss and a conditional GAN framework, we have found that the diversity drop still occurs as we scale up the student and teacher models. We leave further investigation of this problem for future work.
\clearpage

\section{Societal Impact}
Our work aims to develop a one-step image synthesis framework, which could significantly improve the accessibility and affordability of generative visual models. By reducing the multi-step synthesis process into a single step, our technology promises to democratize the creation of visual content, enabling a broader range of users to harness the power of generative models for creative expression and innovation. Additionally, by reducing the need for extensive computation during both training and inference stages, our framework also helps decrease electricity usage and CO2 emissions. However, as this technology becomes more accessible, it is crucial to address concerns about potential misuse, especially in areas like sexual harassment and synthetic media manipulation.

Generative visual models have the potential to facilitate the creation of highly convincing deep fake videos and enable sophisticated impersonation techniques, presenting significant challenges for the trustworthiness of online information. Moreover, they can be utilized to generate content that may incite instances of sexual harassment. While our technology boasts compelling advantages regarding efficiency, it is imperative to acknowledge and tackle the potential societal repercussions and ethical dilemmas linked with the widespread integration of generative visual models. \\

\myparagraph{Acknowledgments.} 
We would like to thank Tianwei Yin, Seungwook Kim, and Sungyeon Kim for their valuable feedback and comments. Part of this work was done while Minguk Kang was an intern at Adobe Research. Minguk Kang and Suha Kwak were supported by the NRF grant and IITP grant funded by Ministry of Science and ICT, Korea (NRF-2021R1A2C3012728, AI Graduate School (POSTECH): RS-2019-II191906). Jaesik Park was supported by the IITP grant funded by the Korea government (MSIT) (AI Graduate School (SNU): RS-2021-II211343 and AI Innovation Hub: RS-2021-II212068). Jun-Yan Zhu was supported by the Packard Fellowship.

\balance
{\small
\bibliographystyle{splncs04}
\bibliography{paper}
}

\clearpage

\appendix

\section*{Appendices}
\addcontentsline{toc}{section}{Appendices}
\renewcommand\thefigure{A\arabic{figure}}
\renewcommand{\thetable}{A\arabic{table}}
\setcounter{figure}{0}
\setcounter{table}{0}

We elaborate on the training details for \OursAcronym{} in Appendix~\ref{appendix:train_eval_details}. Following this, we provide an additional explanation of our proposed E-LatentLPIPS in Appendix~\ref{appendix:elatentlpips}. In Appendix~\ref{appendix:vsGigaGAN}, we offer a quantitative comparison with GigaGAN. Then, we discuss the noise and ODE solution pair dataset in Appendix~\ref{appendix:noise_latent_relation}. Finally, we present additional visuals of \OursAcronym{} and also qualitatively demonstrate that \OursAcronym{} is capable of synthesizing well-aligned and diverse images using a single prompt in Appendix~\ref{appendix:more_visual}.

\section{Training Details}
\label{appendix:train_eval_details}

\subsection{Text-to-Image Synthesis}

\myparagraph{\textbf{Parameterization.}} We distill Stable Diffusion~\cite{stablediffusion1.5, podell2024sdxl} into \OursAcronym{} using the PyTorch framework~\cite{NEURIPS2019_9015}. Throughout our experiments, we utilize the U-Net architecture employed in Stable Diffusion, initializing the U-Net weights with the pre-trained weights of Stable Diffusion. As the Stable Diffusion was originally designed to predict a denoising noise $\eps(\mathbf{x}_{t}, \mathbf{c}, t)$ given a noisy sample~$\mathbf{x}_{t}$, we modify the noise prediction parameterization to the data prediction parameterization using the following equation, though with a slight abuse of notation:
\begin{equation}
    G({\mathbf{x}_{t}, \mathbf{c}, t}) =  \frac{\mathbf{x}_{t} - \sigma_{t}\eps(\mathbf{x}_{t}, \mathbf{c}, t)}{\alpha_{t}},
    \label{eq:eps_param}
\end{equation}
where $\sigma_{t}$ and $\alpha_{t}$ are manually defined diffusion schedule. Since Diffusion2GAN performs noise-to-latent mapping, translating pure Gaussian noise $\mathbf{z}=\mathbf{x}_{T}$ to a target latent $\mathbf{x}=\mathbf{x}_{0}$, the data prediction parameterization mentioned above can be re-written as follows:
\begin{equation}
    G({\mathbf{z}, \mathbf{c}}) =  \frac{\mathbf{z} - \sigma_{T}\eps(\mathbf{z}, \mathbf{c}, T)}{\alpha_{T}}.
    \label{eq:mean_param}
\end{equation}
While it is essential to employ the data prediction parameterization for the generator, as the \OursAcronym{}'s objective is to predict a target latent rather than a denoising noise, we empirically discover that adopting the noise prediction parameterization for the discriminator does not lead to instability issues.

\myparagraph{\textbf{Two-stage \OursAcronym{} training.}} We observe enhanced stability and increased diversity in image generation when employing a two-stage training approach for \OursAcronym{}. In the initial stage, \OursAcronym{} is exclusively trained using the ODE distillation loss. Subsequently, we fine-tune the ODE-distilled one-step generator by incorporating the ODE distillation, conditional GAN, and single-sample R1 losses. Experimentally, we discover that training Diffusion2GAN with a different conditional GAN loss weight typically results in stable convergence. Increasing the weight of the conditional GAN loss component enhances the fidelity of generated images but decreases image diversity. We speculate this occurs because the conditional GAN loss prioritizes realistic image synthesis over accurately learning the original ODE trajectory of the teacher model. Detailed hyperparameters are provided in Table~\ref{table:hyperparameter_details}.

\subsection{Conditional Image Synthesis on CIFAR10-32px}
\myparagraph{\textbf{Consistency Distillation training}}. In Table~\ref{table:cifar10_exp}, we present FID~\cite{Heusel2017GANsTB} of Consistency Distillation~(CD)~\cite{song2023consistency} on CIFAR10~\cite{Krizhevsky2009LearningML}. We implement a conditional version of CD and train it for 150k iterations with a batch size of 512, resulting in $307.2\text{M}=4\times150\text{k}\times512$ number of function evaluations (NFE). Note that the official unconditional CD was trained for 800k iterations to achieve an FID of 3.55, while our conditional CD implementation achieves a nearly identical FID of 3.67 with only 400k training iterations, entailing 819.2M NFE.

\myparagraph{\textbf{ODE distillation training}}. We distill a pre-trained EDM~\cite{karras2022elucidating} on CIFAR10 into a single-step generator only using ODE distillation loss. To create the noise and ODE solution pairs, we simulate the pre-trained EDM 18 times using a Heun sampler~\cite{karras2022elucidating}. When training the ODE distilled generator, we adhere to using the original parameterization of EDM, as the EDM is originally designed to perform data prediction. The hyperparameter details are presented in Table~\ref{table:hyperparameter_details}.

\section{E-LatentLPIPS}
\label{appendix:elatentlpips}
\begin{figure*}[h]
    \centering
    \vspace{-5mm}
    \includegraphics[width=0.8\linewidth]{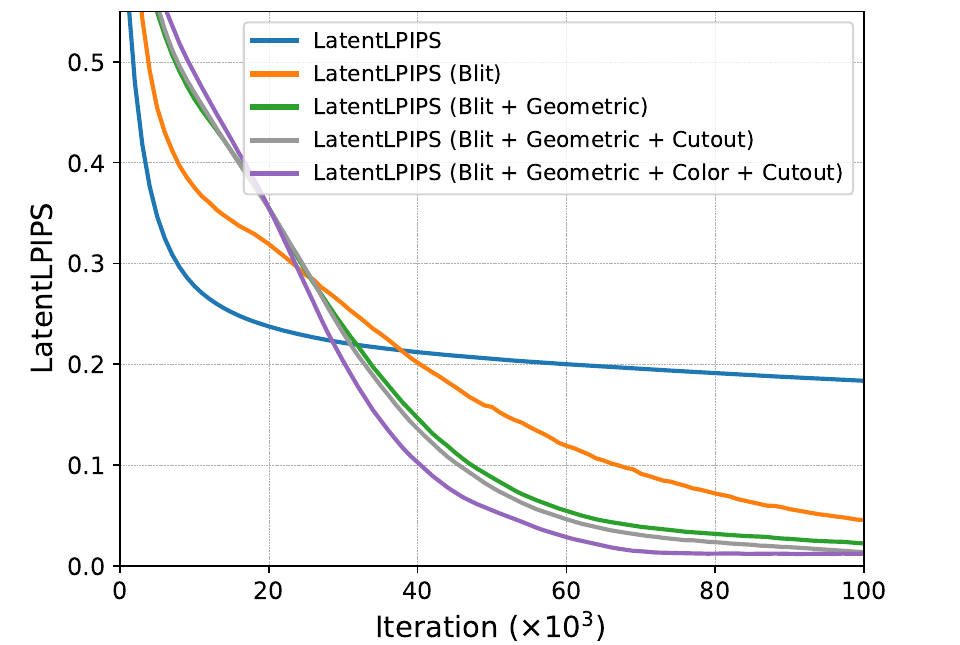}
    \captionsetup{width=1.0\linewidth}
    \captionsetup{singlelinecheck = false, justification=justified}    
    \caption{\textbf{Single sample overfitting experiment}. LatentLPIPS fails to achieve overfitting, even in a single-sample overfitting experiment. However, by applying diverse differentiable augmentations to the inputs of LatentLPIPS, we can successfully reconstruct the target latent. Blit indicates Horizontal flip + 90-degree rotation + integer translation. Geometric indicates isotropic scaling + arbitrary rotation + anisotropic scaling + fractional translation. Color indicates random brightness + random saturation + random contrast. For technical details on the differentiable augmentations, we recommend referring to the papers~\cite{zhao2020differentiable, karras2020training}.}
    \label{fig:elpips_curve}
\end{figure*}

\clearpage

\begin{table*}[t!]
\centering
\caption{ {\bf Hyperparameters for \OursAcronym{} training.} We denote pixel blitting and geometric transformations as bg~\cite{karras2020training} and bg with color transformations as bgc~\cite{karras2020training, zhao2020differentiable}. For additional technical details, please refer to the original papers: LPIPS~\cite{zhang2018unreasonable}, Cutout~\cite{devries2017improved}, Non-saturation loss~\cite{Goodfellow2014GAN}, Adam optimizer~\cite{Kingma2015AdamAM}, RAdam optimizer~\cite{liu2019variance}, EDM~\cite{karras2022elucidating}, SD 1.5~\cite{rombach2022high}, SDXL-Base-1.0~\cite{podell2024sdxl}, Noise augment before $D$~\cite{kang2023scaling, xu2023ufogen}, Heun sampler~\cite{karras2022elucidating}, and DDIM sampler~\cite{song2021denoising}. $\text{E-LatentLPIPS}^{*}$ refers to the ensemble of E-LatentLPIPS with vanilla LatentLPIPS.}
\resizebox{1.0\textwidth}{!}
{
\begin{tabular}{lcccc}
\cmidrule[1.0pt]{1-5}
Model hyperparameters &~~CIFAR10 32px~~&~~SD-CFG-3 64px~~&~~SD-CFG-8 64px~~&~~SDXL-CFG-7 128px~~\\
\cmidrule[1.0pt]{1-5}
$\vb{z}$ dimension & 3$\times$32$\times$32& 4$\times$64$\times$64 & 4$\times$64$\times$64 & 4$\times$128$\times$128\\
$\vb{x}$ dimension & 3$\times$32$\times$32& 4$\times$64$\times$64 & 4$\times$64$\times$64 & 4$\times$128$\times$128\\
$\mathcal{L}_{\text{distill}}^{\text{ODE}}$ loss type & LPIPS &~$\text{E-LatentLPIPS}^{*}$~&~E-LatentLPIPS~&~E-LatentLPIPS~\\
E-LatentLPIPS augmentation & - & bg + cutout & bgc + cutout & bgc + cutout \\
$\mathcal{L}_{\text{distill}}^{\text{ODE}}$ loss strength & 1.0 & 1.0 & 1.0 & 1.0 \\
$\mathcal{L}_\text{GAN}$ loss type & - &~Non-saturation~&~Non-saturation~&~Non-saturation~\\
$\mathcal{L}_\text{GAN}$ loss strength & - & 0.25 & 0.25 & 1.0 \\
Single-sample R1 strength & - & 0.01 & 0.01 & -\\
Single-sample R1 interval & - & 16 & 16 & -\\
Mix-and-match augmentation & False & True & True & True\\
Optimizer & RAdam & Adam & Adam & Adam\\
Batch size & 512 & 256 & 2048$\rightarrow$1024 & 1024$\rightarrow$512\\
Accumulation & 1 & 1 & 1 & 1 \\
$G$ learning rate & 4e-4 & 1e-4$\rightarrow$1e-5 & 1e-4$\rightarrow$1e-5 & 1e-4$\rightarrow$1e-5\\
$G$ $\beta_{1}$ for Adam & 0.9 & 0.9 & 0.9 & 0.9\\
$G$ $\beta_{2}$ for Adam & 0.999 & 0.999 & 0.999 & 0.999\\
$D$ learning rate & - & 1e-4$\rightarrow$1e-5 & 1e-4$\rightarrow$1e-5 & 1e-4$\rightarrow$1e-5\\
$D$ $\beta_{1}$ for Adam & 0.9 & 0.9 & 0.9 & 0.0\\
$D$ $\beta_{2}$ for Adam & 0.999 & 0.999 & 0.999 & 0.99\\
Weight decay strength & 0.0 & 1e-2 & 1e-2 & 1e-2 \\
Weight decay strength on attention & 0.0 & 1e-5 & 1e-5 & 1e-5\\
Dropout rate & 0.1 & 0.0 & 0.0 & 0.0 \\
$\#$ $D$ updates per $G$ update & - & 1 & 1 & 1\\
$G$ ema start & 20k & 4k & 4k & 4k \\
$G$ ema beta & 0.9999 & 0.9999 & 0.9999 & 0.9999\\
Precision & bfloat16 & bfloat16 & bfloat16 & bfloat16\\
$G$ backbone & EDM & SD 1.5 & SD 1.5 & SDXL-Base-1.0 \\
$D$ backbone & - & SD 1.5 & SD 1.5 & SDXL-Base-1.0 \\
Multi-scale training  & - & True & True & True\\
Noise augment before $D$  & False & False & False & True\\
\cmidrule[1.0pt]{1-5}
Training specifications &~~CIFAR10 32px~~&~~SD-CFG-3 64px~~&~~SD-CFG-8 64px~~&~~SDXL-CFG-7 128px~~\\
\cmidrule[1.0pt]{1-5}
Diffusion generator & EDM & SD 1.5 & SD 1.5 & SDXL-Base-1.0\\
Numerical solver & Heun & DDIM & DDIM & DDIM \\
Denoising steps & 18 & 50 & 50 & 50 \\
$\#$ ODE pairs & 1.0M & 3.0M & 12.0M & 8.0M \\
NFE for dataset generation & 35.0M & 150.0M & 600.0M  & 400.0M\\
\cmidrule[1.0pt]{1-5}
Training specifications &~~CIFAR10 32px~~&~~SD-CFG-3 64px~~&~~SD-CFG-8 64px~~&~~SDXL-CFG-7 128px~~\\
\cmidrule[1.0pt]{1-5}
$G$ Model size & 61.5M & 859.5M & 859.5M & 2567.5M\\
$D$ Model size & - & 859.6M & 859.6M & 2567.7M\\
First stage iterations & 150k & 150k & 50k & 20k\\
Second stage iterations & - & 10k & 10k & 30k\\
NFE for training & 76.8M & 51.2M & 153.6M & 97.3M\\
GPU type & A100 & A100-80GB & A100-80GB & A100-80GB\\
$\#$ GPUs for training & 8 & 16 & 64 & 128 \\
GPU days & 6.0  & 43.6 & 119.2 & 356.8\\
FID & 3.16 &9.29 & 13.39 & 25.49\\
\cmidrule[1.0pt]{1-5}
\end{tabular}}
\label{table:hyperparameter_details}
\end{table*}
\clearpage

\subsection{Toy Experiment}
We conducted a single image reconstruction experiment to study how LatentLPIPS behaves. Beginning with a 512-pixel target image, denoted as $\mathbf{I}_{\text{target}} \in \mathbb{R}^{3\times512\times512}$, we utilized the VAE encoder of Stable Diffusion to obtain its latent vector, resulting in $\mathbf{x}_{\text{target}} = \text{Encode}^{1/8\times}(\mathbf{I}_{\text{target}}) \in \mathbb{R}^{4\times64\times64}$. Subsequently, we randomly initialized a trainable latent vector $\mathbf{x}_{\text{source}}$ with the same dimensions as $\mathbf{x}_{\text{target}}$. The objective of this experiment is to determine whether LatentLPIPS can achieve a latent vector $\mathbf{x}_{\text{source}}$ that precisely reconstructs $\mathbf{x}_{\text{target}}$ using the following LatentLPIPS objective and a gradient-based optimizer:
\begin{equation}
    \dist_\text{LatentLPIPS}(\image_{\text{target}}, \image_{\text{source}}) = \ell(F(\image_{\text{target}}), F(\image_{\text{source}})),
    \label{eq:elatentlpips_toy}
\end{equation}
where $F$ is a VGG network trained in the latent space of Stable Diffusion, and $\ell(\cdot, \cdot)$ is a distance metric. While a well-designed single sample overfitting is typically considered feasible, our analysis suggests that LatentLPIPS struggles with optimization, resulting in a high loss value, as shown in Figure~\ref{fig:elpips_curve}. Moreover, we observed systematic wavy and patchy artifacts in the reconstructed image decoded by the source latent. We hypothesize that this limitation arises from a suboptimal loss landscape created by the latent version of the VGG network.

Inspired by E-LPIPS~\cite{kettunen2019lpips} and the observation that only a portion of the region has been successfully reconstructed using the source latent, we apply geometric augmentations and cutout~\cite{devries2017improved} to both the source and target latents. To ensure differentiability for backpropagation, we employ off-the-shelf differentiable augmentations ~\cite{zhao2020differentiable, karras2020training}. Upon introducing these augmentations, we notice improved convergence of LatentLPIPS, suggesting that the poor optimization can be alleviated by applying an appropriate combination of differentiable augmentations. Through toy experiments, we have confirmed that LatentLPIPS converges faster and better as we introduce more augmentations, including augmentations related to color (random brightness, saturation, and contrast), as shown in Figure~\ref{fig:elpips_curve}.

In text-to-image experiments, we found that the combination of generic geometric transformations and cutout achieves the best FID on the SD-CFG-3 dataset, while additionally using the color-related augmentations proves beneficial for the SD-CFG-8 and SDXL-CFG-7 datasets. Furthermore, we discovered that on SD-CFG-3, \OursAcronym{} achieves better FID when E-LatentLPIPS is combined with vanilla LatentLPIPS.

\subsection{Perceptual Score of LatentLPIPS vs. LPIPS}

In Section~\ref{sec:generator}, we described learning LatentLPIPS, following the procedure from LPIPS~\cite{zhang2018unreasonable}. This involves training an ImageNet~\cite{Deng2009ImageNetAL} classifier and then tuning it to perceptual scores.

In Table~\ref{tab:imagenet}, we present ImageNet classification accuracies. The LPIPS network uses VGG16~\cite{simonyan2014very} as a backbone, which achieves $71.59\%$ accuracy. We note that a batch-norm version of the backbone achieves $73.36\%$. The ImageNet classification score on latent codes drops to $64.25\%$, while the batch-norm variant recovers some performance on $68.26\%$. We found the batch-norm variant trains more stably. We followed the default PyTorch training code and parameters~\url{https://github.com/pytorch/examples/blob/main/imagenet/main.py}, but discovered that we had to reduce the initial learning rate for the non-batch-norm variant. We selected the batch-norm version to form the basis of LatentLPIPS. While the ImageNet classification scores are lower, they are competitive in terms of perceptual quality measurement. More importantly, as noted in the original LPIPS work, ImageNet classification scores do not necessarily correlate with perceptual quality -- ImageNet classification is merely a pretext task to yield a representation with high perceptual quality.

In Table~\ref{tab:perceptual_scores}, we show the perceptual scores on the Berkeley-Adobe Perceptual Patch Similarity (BAPPS) dataset~\cite{zhang2018unreasonable}. The dataset provides different types of perturbations, ``traditional'' hand-crafted perturbations, ones from CNN-generated algorithms, and outputs from real algorithms for image reconstruction tasks (colorization, video interpolation, superresolution, and video deblurring). We followed the protocol from LPIPS~\cite{zhang2018unreasonable}, learning a linear calibration on 5 different intermediate layers. Across the different sets, LatentLPIPS achieves similar, sometimes higher scores, as vanilla LPIPS. This indicates that while some details that are advantageous for classification are lost during compression, the perceptually important details are preserved. This result aligns with the goal of designing the latent space~\cite{rombach2022high} in the first place. In conclusion, our LatentLPIPS is able to capture a representation that aligns with human perception, at similar performance to vanilla LPIPS, while enabling faster computation. Please note that extra training for LatentLPIPS was performed to distill SDXL-Base-1.0 into \OursAcronym{} because Stable Diffusion 1.5 and SDXL-Base-1.0 do not share the same latent space.

\begin{table}[h]
    \centering
    \caption{\textbf{ImageNet classification scores.} The backbone networks in $^{*}$ are used for LPIPS~\cite{zhang2018unreasonable} $\&$ LatentLPIPS calculations. ImageNet accuracy on the Latent code is lower than on pixels, as information is lost during compression. However, ImageNet classification is merely a proxy task for achieving a strong representation to align with human perception. The perceptual scores in Table~\ref{tab:perceptual_scores} are competitive, indicating perceptual information is retained.}
    \vspace{-2mm}
    \begin{tabular}{lcc}
        \cmidrule[1.0pt]{1-3}
        Perceptual metric~&~VGG16~&~VGG-bn~\\ 
        \cmidrule[1.0pt]{1-3}
        Pixels &~\textbf{71.59}$^{*}$~&~\textbf{73.36}~\\
        Latent &64.25~&~~68.26$^{*}$~\\
        \cmidrule[1.0pt]{1-3}
    \end{tabular}
    \label{tab:imagenet}
\end{table}
\vspace{-12mm}
\begin{table}[h]
    \centering
    \caption{\textbf{Perceptual scores.} LatentLPIPS achieves similar and sometimes higher perceptual scores than vanilla LPIPS~\cite{zhang2018unreasonable} on the BAPPS dataset.}
    \vspace{-2mm}
    \begin{tabular}{lccc}
        \cmidrule[1.0pt]{1-4}
        Perceptual metric~~&~Traditional~&~CNN~&~Real~\\ 
        \cmidrule[1.0pt]{1-4}
        LPIPS~\cite{zhang2018unreasonable} &~73.36~&~\textbf{82.20}~&~\textbf{63.23}~\\
        LatentLPIPS &~\textbf{74.29}~&~81.99~&~63.21~\\
        \cmidrule[1.0pt]{1-4}
    \end{tabular}
    \label{tab:perceptual_scores}
\end{table}
\clearpage

\section{Quantitative Comparison with GigaGAN}
\label{appendix:vsGigaGAN}
\vspace{-7mm}
\begin{table}[h!]
\centering
\caption{{\bf Comparison to text-to-image GigaGAN generator on COCO2014}. While our \OursAcronym{} model shows a slightly higher FID~\cite{Heusel2017GANsTB} compared to GigaGAN~\cite{kang2023scaling}, it exhibits a higher recall value~\cite{Kynknniemi2019ImprovedPA}, indicating that \OursAcronym{} can generate more diverse images than GigaGAN.}
\vspace{-2mm}
\resizebox{0.95\textwidth}{!}{
\begin{tabular}{lcccccc}
\cmidrule[1.0pt]{1-6}
 model~&~FID-30k~($\downarrow$)~ & ~CLIP-30k~($\uparrow$)~ & ~Precision~($\uparrow$)~ & ~Recall~($\uparrow$)~ & ~A100 days~\\
\cmidrule[1.0pt]{1-6}
GigaGAN~\cite{kang2023scaling} & \textbf{9.09} & \textbf{0.32} & \textbf{0.74} & 0.60 & 4783.0 \\
\textbf{\OursAcronym{}} & 9.29 & 0.31 & 0.74 & \textbf{0.64} & \textbf{43.6} \\
\cmidrule[1.0pt]{1-6}
\end{tabular}}
\vspace{-5mm}
\label{tab:coco2014_gigagan}
\end{table}
We compare Diffusion2GAN with GigaGAN~\cite{kang2023scaling} using additional metrics, including Clip-score~\cite{hessel2021clipscore} and Precision $\&$ Recall~\cite{Kynknniemi2019ImprovedPA}. We utilize the officially provided GigaGAN samples~\cite{gigagan2023eval} to compute these metrics. As shown in Table~\ref{tab:coco2014_gigagan}, \OursAcronym{} achieves a higher recall than GigaGAN, suggesting that \OursAcronym{} suffers less from diversity collapse than GigaGAN. Despite slightly worse FID and Clip-score, \OursAcronym{} achieves almost comparable performance while using only about 1$\%$ of the compute resources.

\section{Discussion on Noise and ODE Solution Pair Dataset}
\label{appendix:noise_latent_relation}
\vspace{-1mm}
In this paper, we create noise-image (latent) pairs using a pre-trained diffusion model and a deterministic sampler. This prompts fundamental questions: should these pairs strictly adhere to a one-to-one correspondence, and can they be randomly re-paired while still maintaining this correspondence? To explore these questions, we generate noise-image pairs using a stochastic sampler. Specifically, we utilize a pre-trained EDM~\cite{karras2022elucidating} and generate 50k noise-image pairs using an EDM's stochastic sampler. Subsequently, we train a one-step model using ODE distillation loss with LPIPS, as explained in Section~\ref{sec:generator}. However, the one-step model with stochastic pairs cannot minimize the ODE distillation loss, resulting in an FID over 200. This phenomenon also occurs when we randomly re-wire 50k deterministic noise-image pairs without replacement. This result contradicts our earlier findings, where a model trained using ODE distillation loss achieved an FID score of 8.51 using 50k diffusion-simulated deterministic noise-image pairs, as presented in Table~\ref{table:cifar10_exp}. These results suggest that for effective ODE distillation, noise-image pairs should be deterministically generated and inherit a specific relationship formed by simulating the ODE of a pre-trained diffusion model.

\section{More Visual Results}
\label{appendix:more_visual}
\begin{figure*}[htp!]
    \centering
    \includegraphics[width=0.94\linewidth]{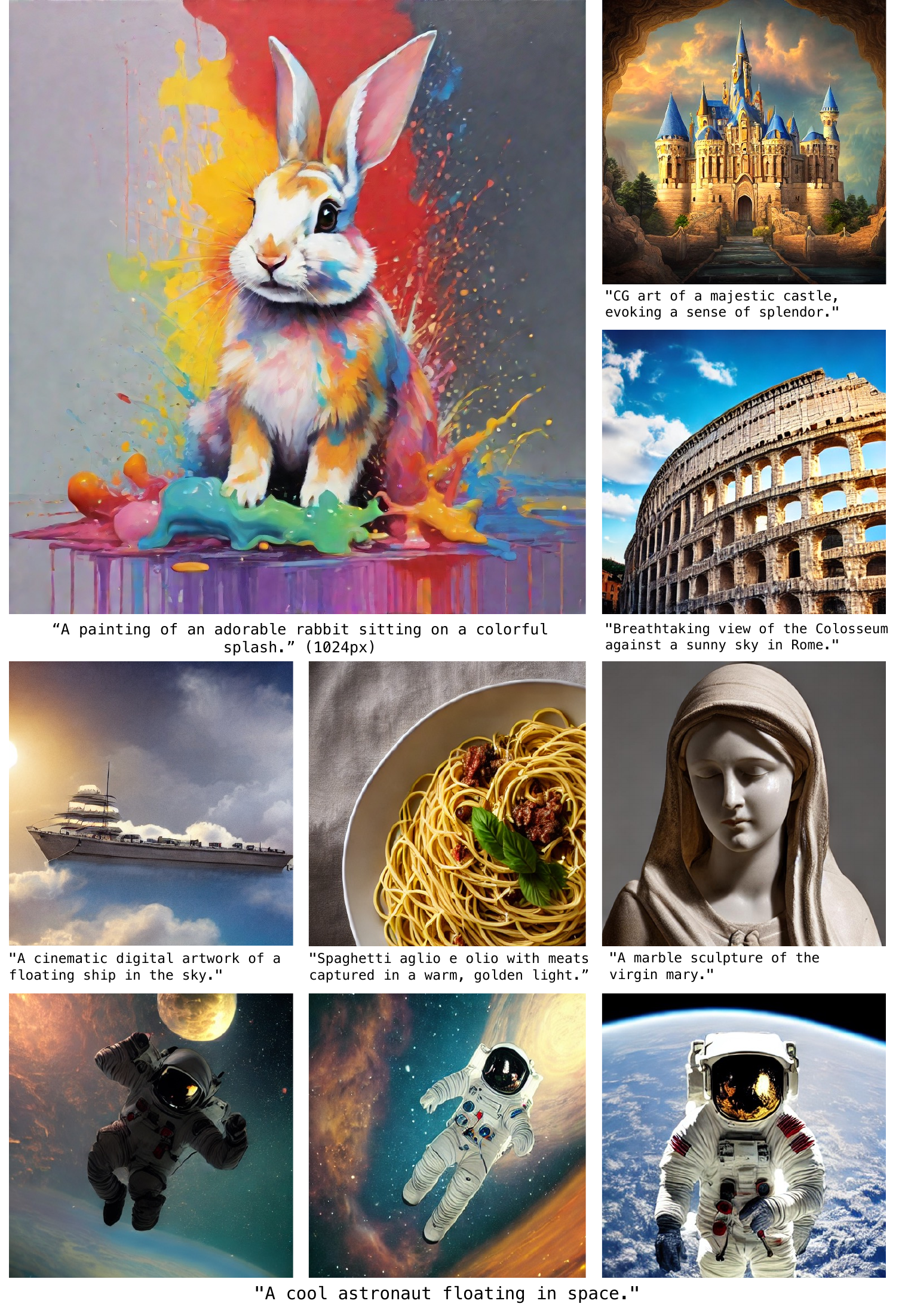}
    \vspace{-3mm}
    \captionsetup{width=1.0\linewidth}
    \captionsetup{singlelinecheck = false, justification=justified}    
    \caption{High-quality generated images using our one-step \OursAcronym{} framework. Our model can synthesize a 512px/1024px image at an interactive speed of 0.09/0.16 seconds on an A100 GPU, while the teacher model, Stable Diffusion 1.5~\cite{stablediffusion1.5}/SDXL~\cite{podell2024sdxl}, produces an image in 2.59/5.60 seconds using 50 steps of the DDIM~\cite{song2021denoising}.}
    \label{fig:text2image_results_appendix}
\end{figure*}

\begin{figure*}[htp!]
    \centering
    \includegraphics[width=0.99\linewidth]{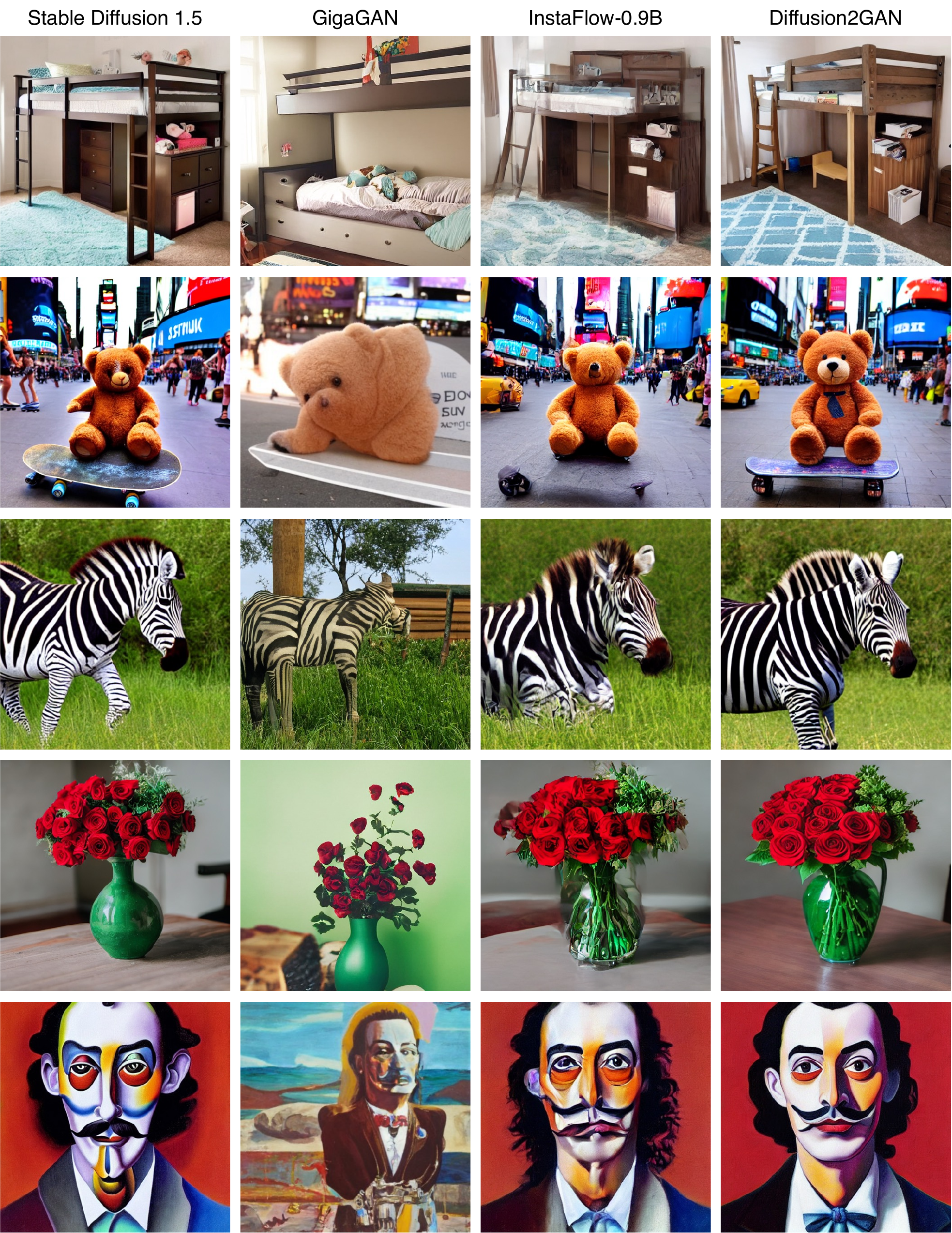}
    \captionsetup{width=1.0\linewidth}
    \captionsetup{singlelinecheck = false, justification=justified}    
    \caption{Visual comparison to Stable Diffusion 1.5~\cite{stablediffusion1.5} with a guidance scale of 8~\cite{ho2022classifier} and selected one-step generators, GigaGAN~\cite{kang2023scaling}, InstaFlow-0.9B~\cite{liu2023insta}, and Diffusion2GAN trained on SD-CFG-8. We observe that Diffusion2GAN produces more realistic images compared to GigaGAN and InstaFlow-0.9B, while maintaining comparable visual quality with Stable Diffusion 1.5.}
    \label{fig:text2image_comparison0}
\end{figure*}

\begin{figure*}[htp!]
    \centering
    \includegraphics[width=0.99\linewidth]{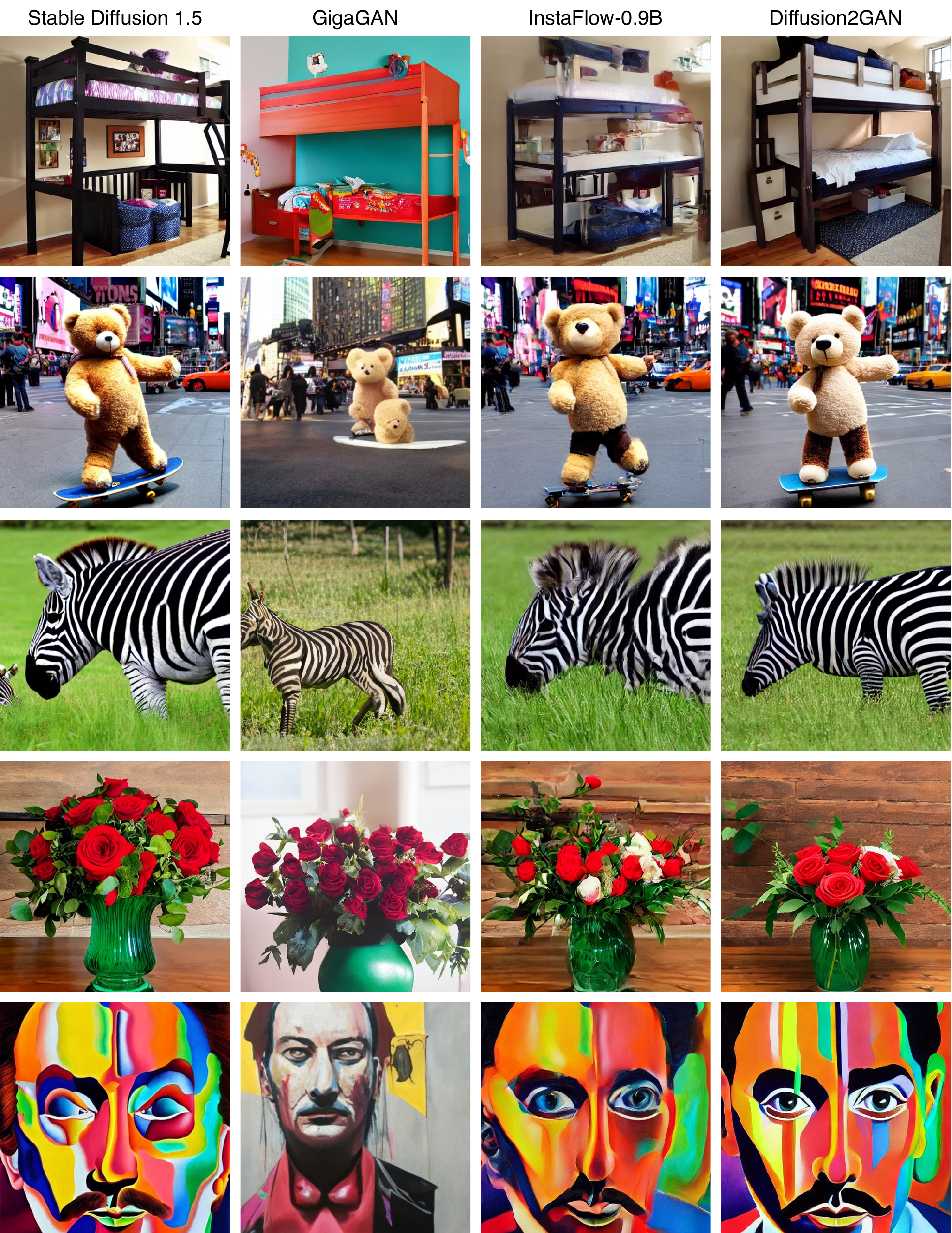}
    \captionsetup{width=1.0\linewidth}
    \captionsetup{singlelinecheck = false, justification=justified}    
    \caption{Visual comparison to Stable Diffusion 1.5~\cite{stablediffusion1.5} with a guidance scale of 8~\cite{ho2022classifier} and selected one-step generators, GigaGAN~\cite{kang2023scaling}, InstaFlow-0.9B~\cite{liu2023insta}, and Diffusion2GAN trained on SD-CFG-8. We observe that Diffusion2GAN produces more realistic images compared to GigaGAN and InstaFlow-0.9B, while maintaining comparable visual quality with Stable Diffusion 1.5.}
    \label{fig:text2image_comparison1}
\end{figure*}

\begin{figure*}[htp!]
    \centering
    \includegraphics[width=0.99\linewidth]{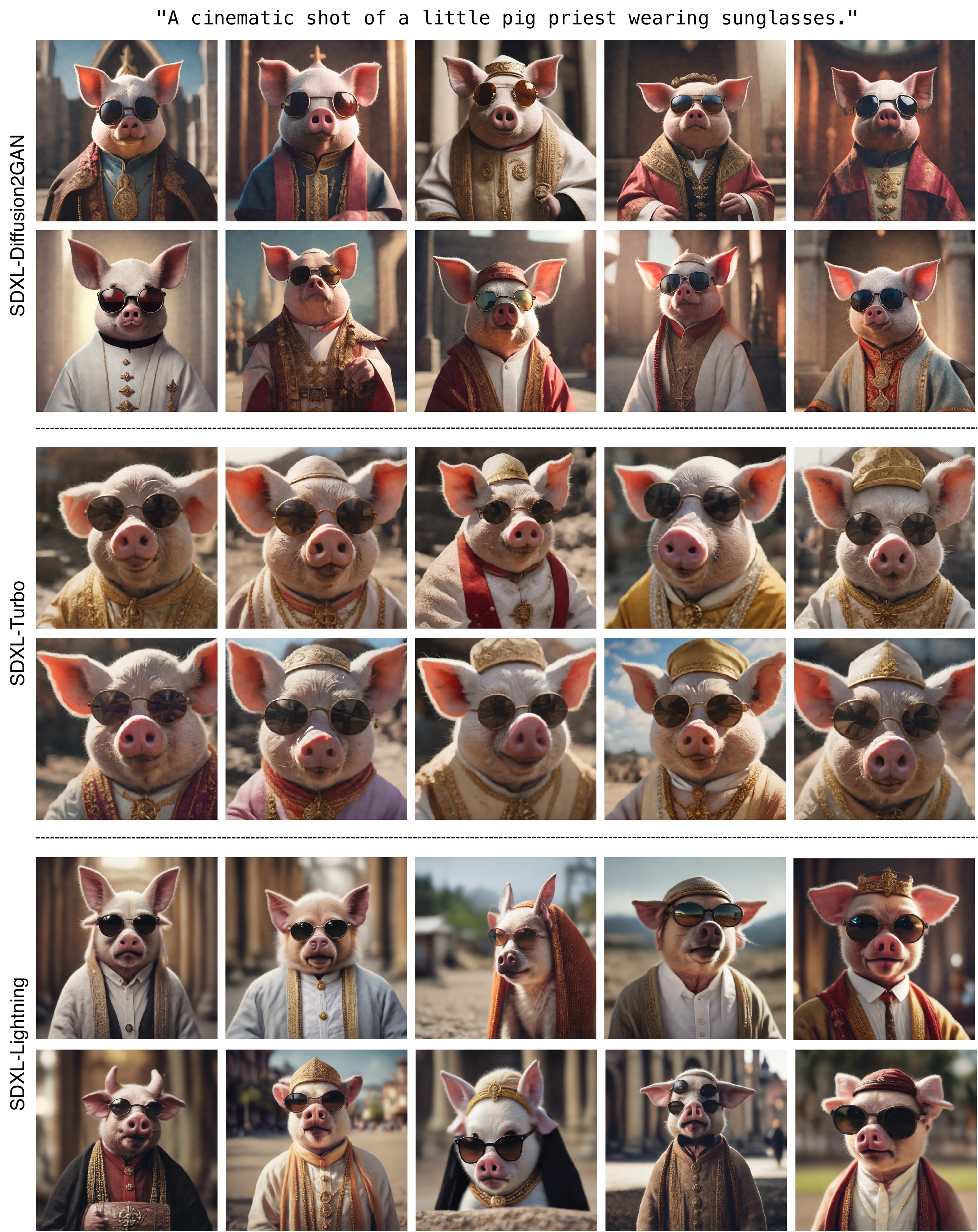}
    \captionsetup{width=1.0\linewidth}
    \captionsetup{singlelinecheck = false, justification=justified}    
    \caption{\textbf{Diversity of generated images from one-step diffusion distillation models}. By altering the random seed used for sampling Gaussian noises, \OursAcronym{} can generate diverse images that closely align with the provided prompt.}
    \label{fig:diversity2}
\end{figure*}

\begin{figure*}[htp!]
    \centering
    \includegraphics[width=0.99\linewidth]{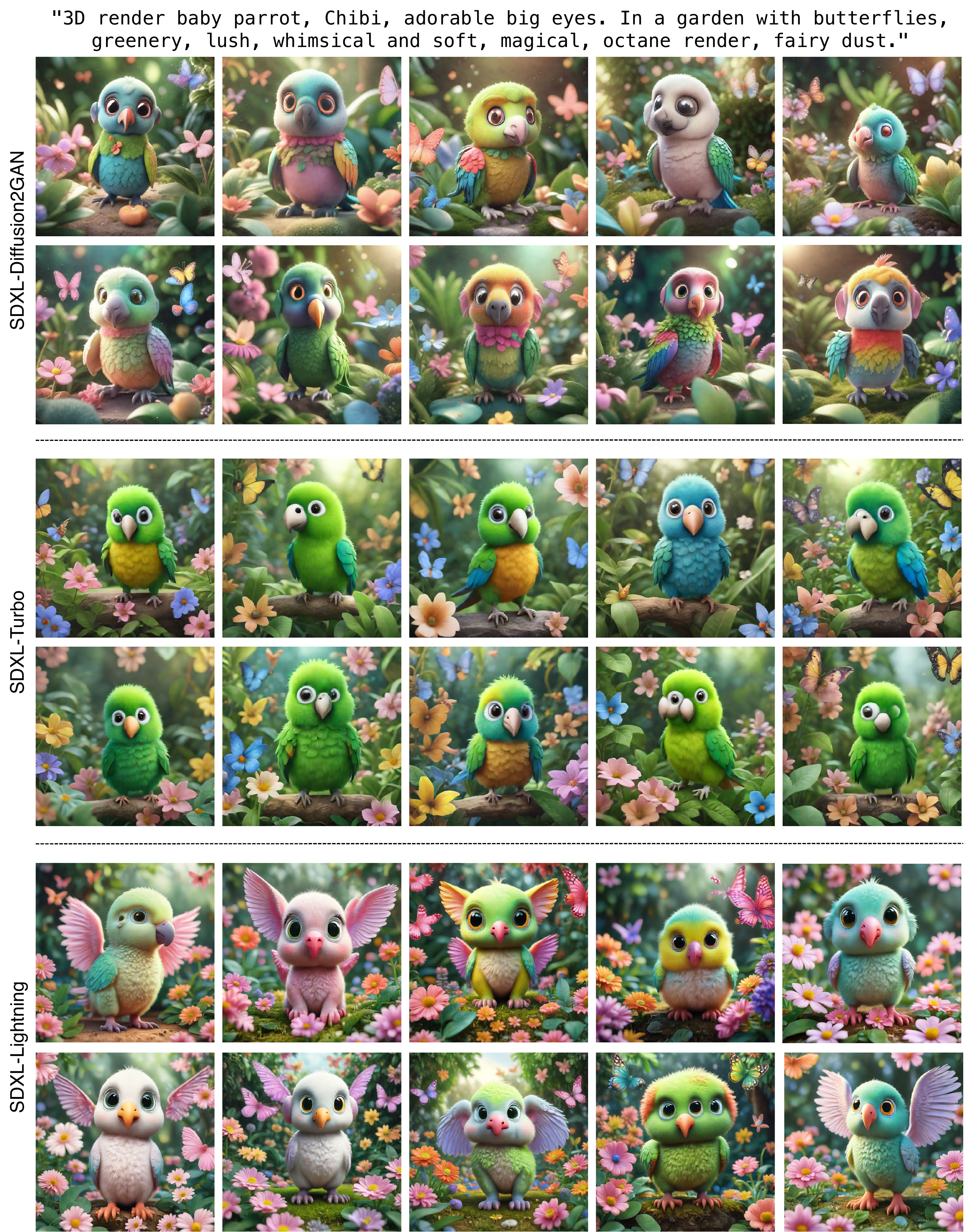}
    \captionsetup{width=1.0\linewidth}
    \captionsetup{singlelinecheck = false, justification=justified}    
    \caption{\textbf{Diversity of generated images from one-step diffusion distillation models}. By altering the random seed used for sampling Gaussian noises, \OursAcronym{} can generate diverse images that closely align with the provided prompt.}
    \label{fig:diversity1}
\end{figure*}

\vspace{-1mm}
We provide additional visuals from Diffusion2GAN in Figure~\ref{fig:text2image_results_appendix}. We also present additional visual comparison between Stable Diffusion 1.5~\cite{stablediffusion1.5}, GigaGAN~\cite{kang2023scaling}, InstaFlow-0.9B~\cite{liu2023insta}, and our Diffusion2GAN using COCO2014 prompts in Figures~\ref{fig:text2image_comparison0} and~\ref{fig:text2image_comparison1}. Furthermore, we demonstrate that SDXL-Diffusion2GAN can generate diverse images from a single prompt while maintaining better text-to-image alignment compared to SDXL-Turbo and SDXL-Lightning in Figures~\ref{fig:diversity2} and~\ref{fig:diversity1}.

\clearpage
\clearpage
\clearpage
\clearpage
\clearpage

\end{document}